\documentclass[final]{cvpr}

\usepackage{times}
\usepackage{graphicx}
\usepackage{amsmath}
\usepackage{amssymb}
\usepackage{booktabs}
\usepackage[utf8]{inputenc} 
\usepackage[T1]{fontenc}    
\usepackage{amsfonts}       
\usepackage{nicefrac}       
\usepackage{microtype}      
\usepackage{xcolor}         
\usepackage{graphicx}
\usepackage{array}
\usepackage{xspace}
\usepackage{colortbl} 
\usepackage{newtxtext} 
\usepackage{relsize}

\usepackage{framed}
\usepackage{enumerate}
\usepackage{enumitem}
\usepackage{caption}
\usepackage{multirow}
\usepackage{multicol}
\usepackage{subfig}
\usepackage{floatrow}
\usepackage{enumitem}
\usepackage{array}
\usepackage{calc}
\usepackage{setspace}

\usepackage[pagebackref=true,breaklinks=true,colorlinks,bookmarks=false]{hyperref}

\renewcommand\@{@}

\renewcommand{\spadesuit}[0]{\text{\smash{\raisebox{-1pt}{\includegraphics[height=8pt]{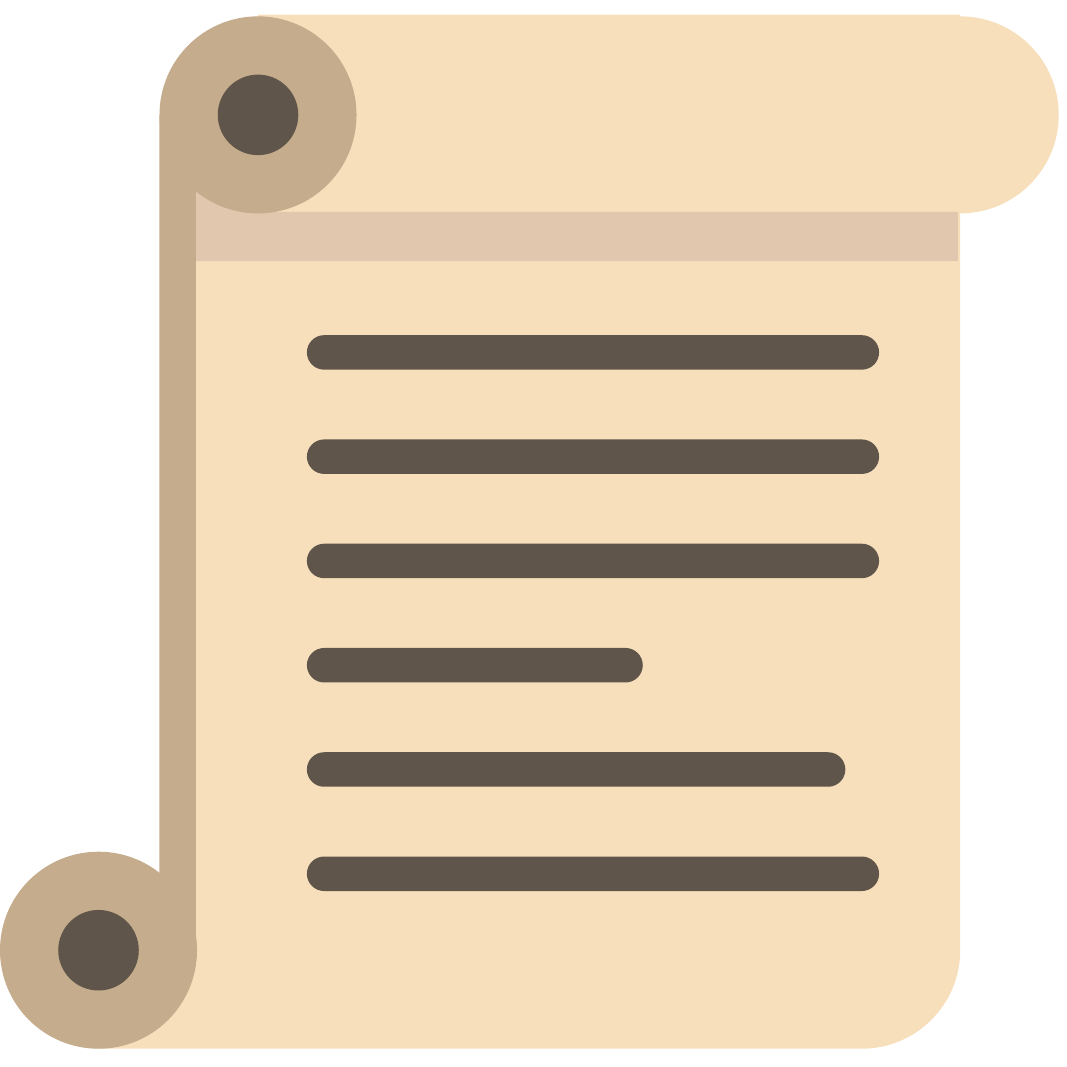}}}}}
\renewcommand{\heartsuit}[0]{\text{\smash{\raisebox{-1pt}{\includegraphics[height=8pt]{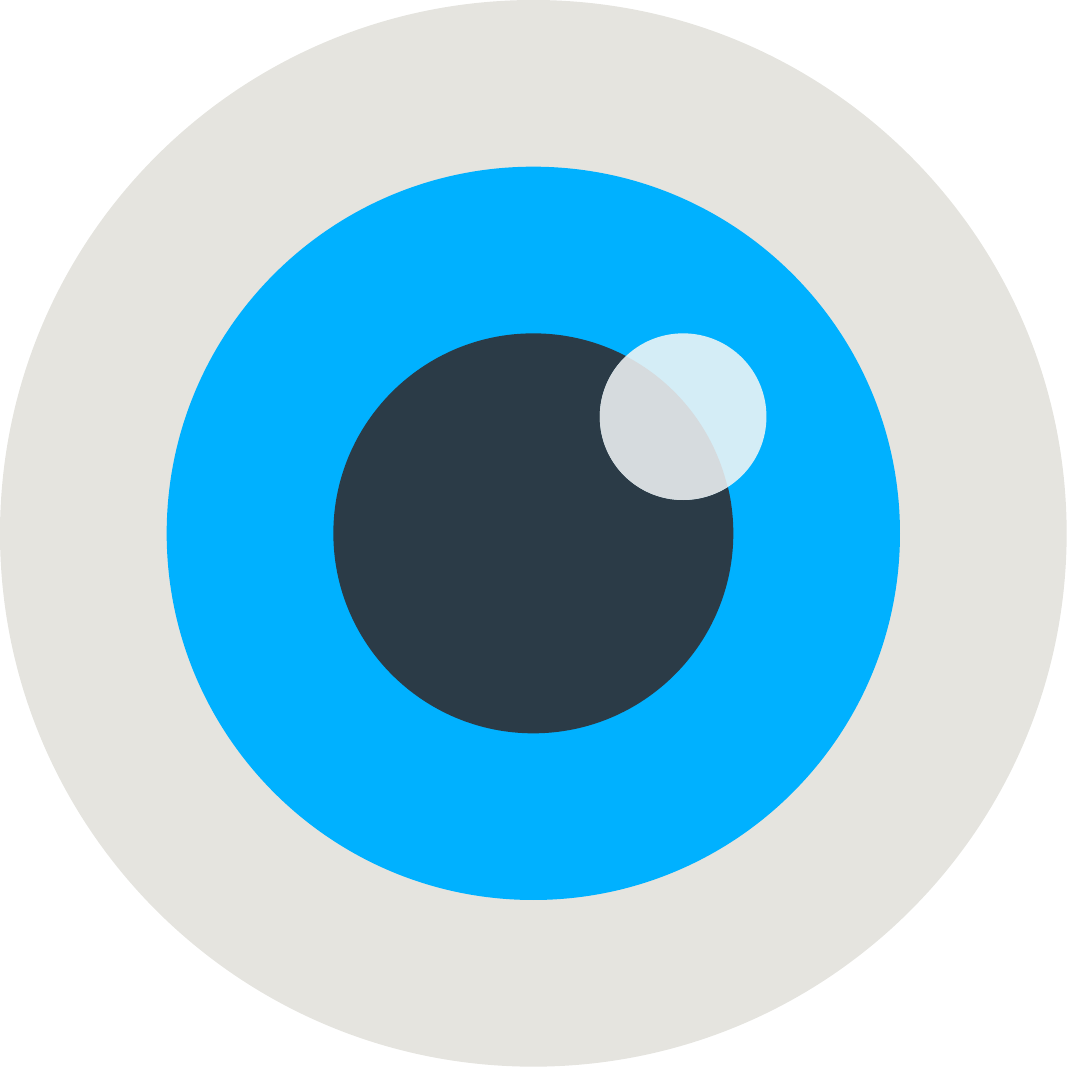}}}}}
\renewcommand{\diamondsuit}[0]{\text{\smash{\raisebox{-1pt}{\includegraphics[height=8pt]{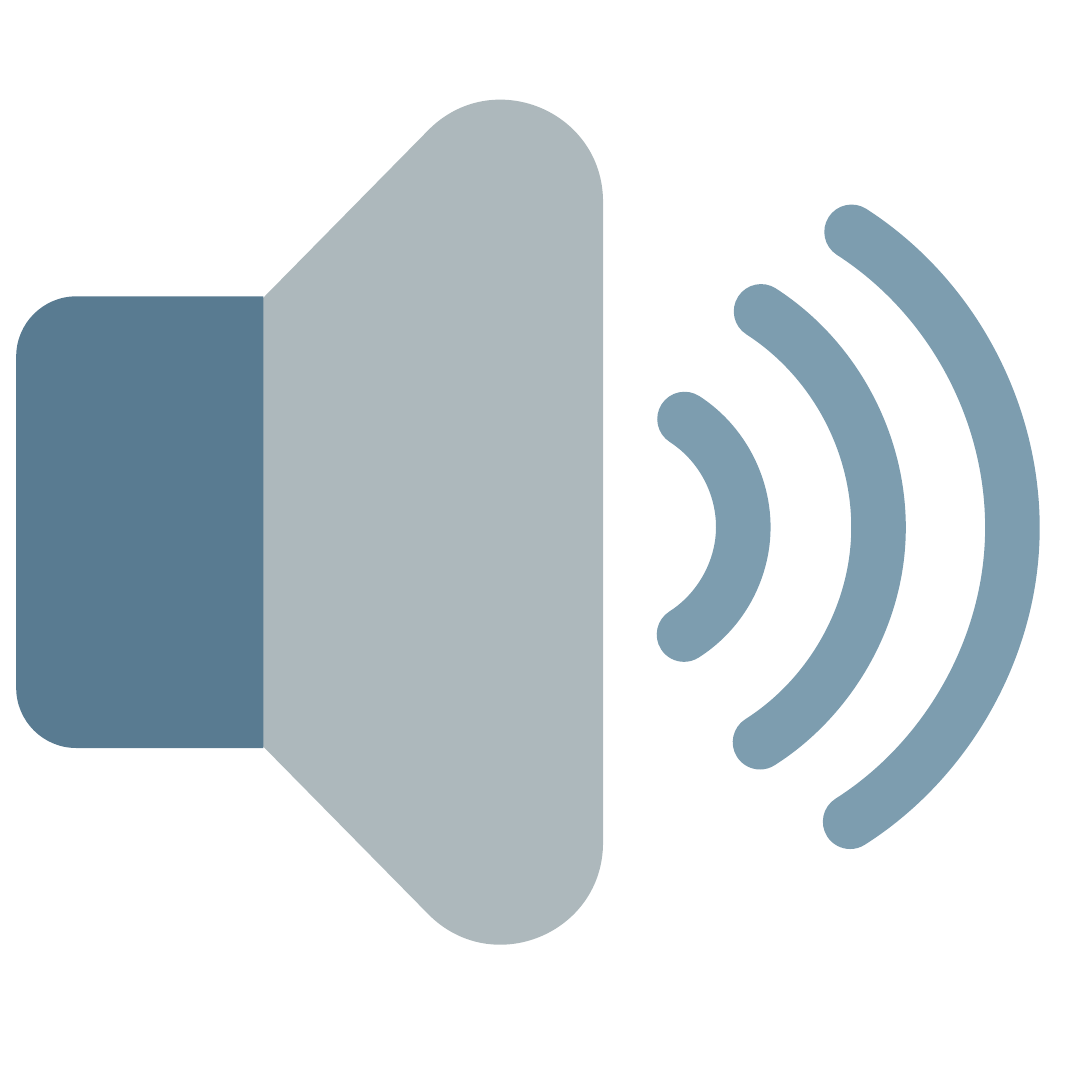}}}}}

\newcommand\merlottitlefont[1]{\smash{{\usefont{T1}{cinzeldecorativebold}{m}{n}#1}}}
\newcommand\merlotfont[1]{\smash{{\usefont{T1}{cinzeldecorative}{m}{n}#1}}}

\newcommand{\wineglassemoji}[0]{\smash{\raisebox{-2pt}{\includegraphics[height=\heightof{\larger\usefont{T1}{cinzeldecorativebold}{m}{n} R}]{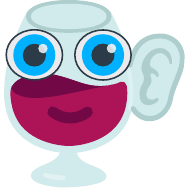}}}}
\newcommand{\wineglassemojihalfwidth}[0]{\smash{\raisebox{-2pt}{\includegraphics[height=0.13in]{merlotreservelogo.pdf}}}}

\newcommand{\modelnamemedium}{\wineglassemoji\merlotfont{MERLOT Reserve}\xspace}

\newcommand{\modelname}{\wineglassemoji\merlotfont{Reserve}\xspace}
\newcommand{\merlottitle}{\merlottitlefont{MERLOT Reserve}}
\newcommand{\modelnamehalfwidth}{\wineglassemojihalfwidth\merlotfont{Reserve}\xspace}

\newcommand{\modelnamelong}{\merlottitlefont{M}ultimodal \merlottitlefont{E}vent \merlottitlefont{R}epresentation \merlottitlefont{L}earning \merlottitlefont{O}ver \merlottitlefont{T}ime, with \merlottitlefont{Re}-entrant \merlottitlefont{S}up\merlottitlefont{erv}ision of \merlottitlefont{E}vents\xspace}

\newcommand{\basemodel}{\wineglassemoji\smash{\merlotfont{Reserve}-\maskblockyfont{B}}\xspace}
\newcommand{\largemodel}{\wineglassemoji\smash{\merlotfont{Reserve}-\maskblockyfont{L}}\xspace}
\newcommand{\basemodelhalfwidth}{\wineglassemojihalfwidth\merlotfont{Reserve}-\maskblockyfont{B}\xspace}
\newcommand{\largemodelhalfwidth}{\wineglassemojihalfwidth\merlotfont{Reserve}-\maskblockyfont{L}\xspace}

\newcommand{\mparagraph}[1]{\textbf{{#1}}}

\newcommand\maskblockyfont[1]{{\usefont{T1}{robotomono}{m}{n}#1}}
\newcommand\masktoken{{\smaller\maskblockyfont{MASK}}}
\newcommand\maskaudiotoken{{\smaller\maskblockyfont{MASKAUDIO}}}
\newcommand\clstoken{{\smaller\maskblockyfont{CLS}}}

\newfloatcommand{capbtabbox}{table}[][\FBwidth]

\newcommand{\merlotdatasetname}[0]{YT-Temporal-180M}

\newcommand{\datasetname}[0]{YT-Temporal-1B}
\newcommand{\vttota}[0]{{\tt\small VT}$\rightarrow${\tt\small TA}}
\newcommand{\vtatot}[0]{{\tt\small VTA}$\rightarrow${\tt\small T}}

\SetEnumitemKey{rowan}{wide, labelwidth=!,listparindent=0pt, labelindent=0pt,noitemsep,topsep=4pt,parsep=2pt,leftmargin =*,label=\textbf{\alph*.}}
\SetEnumitemKey{rowanroman}{wide, labelwidth=0pt,labelsep=-3pt,listparindent=0pt, labelindent=0pt,noitemsep,topsep=4pt,parsep=2pt,leftmargin=*,label=\textbf{\roman*.}}

\DeclareFontFamily{U}{mathb}{}
\DeclareFontShape{U}{mathb}{m}{n}{
  <-5.5> mathb5
  <5.5-6.5> mathb6
  <6.5-7.5> mathb7
  <7.5-8.5> mathb8
  <8.5-9.5> mathb9
  <9.5-11.5> mathb10
  <11.5-> mathb12
}{}
\DeclareSymbolFont{mathb}{U}{mathb}{m}{n}
\DeclareMathSymbol{\ulsh}{3}{mathb}{"E8}
\DeclareMathSymbol{\ursh}{3}{mathb}{"E9}
\DeclareMathSymbol{\dlsh}{3}{mathb}{"EA}
\DeclareMathSymbol{\drsh}{3}{mathb}{"EB}

\newcommand{\websitelink}{{\tt \href{https://rowanzellers.com/merlotreserve}{rowanzellers.com/merlotreserve}}}

\title{\vspace{-4.5mm}\merlottitle:\\ Neural Script Knowledge through Vision and Language and Sound\vspace{-4mm}}
\author{
  Rowan Zellers$^\spadesuit$ \: \: 
  Jiasen Lu$^\heartsuit$ \: \: 
  Ximing Lu$^{\spadesuit\heartsuit}$ \: \: 
  Youngjae Yu$^{\heartsuit}$ \: \: 
  Yanpeng Zhao$^{\diamondsuit}$ \: \: \\[1pt]
  Mohammadreza Salehi$^{\spadesuit}$ \: \:
  Aditya Kusupati$^{\spadesuit}$ \: \: 
  Jack Hessel$^{\heartsuit}$ \: \:
  Ali Farhadi$^{\spadesuit}$ \: \:
  Yejin Choi$^{\spadesuit\heartsuit}$\\[2.5mm]
  $^\spadesuit$Paul G. Allen School of Computer Science \& Engineering, University of Washington \\[-0.25mm]
  $^\heartsuit$Allen Institute for Artificial Intelligence \qquad  $^\diamondsuit$University of Edinburgh\\[-0.25mm]
  \websitelink\vspace*{-2mm}
  }

\begin{document}

\twocolumn[{
\renewcommand\twocolumn[1][]{#1}
\maketitle
\vspace*{-5.5mm}
\centering
\iftoggle{cvprfinal}{\includegraphics[width=\linewidth]{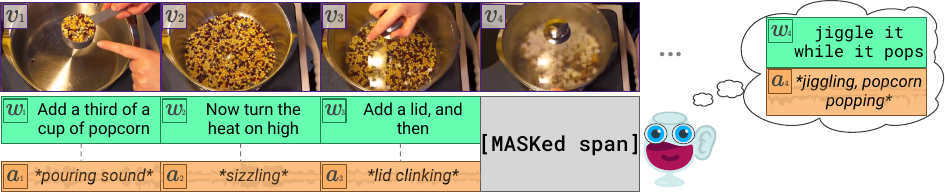}}{\includegraphics[width=\linewidth]{figures/reserve-teaser2.pdf}}%
\vspace{-0.5mm}
\captionof{figure}{\modelnamemedium~learns \emph{multimodal neural script knowledge} representations of video -- jointly reasoning over video frames, text, and audio. Our model is pretrained to predict which snippet of text (and audio) might be hidden by the \masktoken. This task enables it to perform well on a variety of vision-and-language tasks, in both zero-shot and finetuned settings.}
\label{fig:teaser}
\vspace*{5mm}
}]

\maketitle
\begin{abstract}
\vspace*{-3mm}
As humans, we navigate a multimodal world, building a holistic understanding from all our senses.
We introduce \modelnamemedium, a model that represents videos jointly over time -- through a new training objective that learns from audio, subtitles, and video frames.
Given a video, we replace snippets of text and audio with a \masktoken~token; the model learns by choosing the correct masked-out snippet. 
Our objective learns faster than alternatives, and performs well at scale: we pretrain on 20 million YouTube videos.

Empirical results show that \modelnamemedium learns strong multimodal representations. When finetuned, it sets state-of-the-art on Visual Commonsense Reasoning (VCR), TVQA, and Kinetics-600; outperforming prior work by 5\%, 7\%, and 1.5\% respectively. Ablations show that these tasks benefit from audio pretraining -- even VCR, a QA task centered around images (without sound). 
Moreover, our objective enables out-of-the-box prediction, revealing strong multimodal commonsense understanding. In a fully zero-shot setting, our model obtains competitive results on four video tasks, even outperforming supervised approaches on the recently proposed Situated Reasoning (STAR) benchmark. %

We analyze why audio enables better vision-language representations, suggesting significant opportunities for future research. We conclude by discussing ethical and societal implications of multimodal pretraining.
\end{abstract}
\vspace*{-5mm}

\section{Introduction}
\vspace*{-1mm}
The world around us is dynamic. We experience and learn from it using all of our senses, reasoning over them temporally through \emph{multimodal script knowledge} \cite{Schank1975, zellers2021merlot}. Consider Figure~\ref{fig:teaser}, which depicts someone cooking popcorn. From the images and dialogue alone, we might be able to imagine what \emph{sounds} of the scene are: the process might begin with raw kernels scattering in an empty, metallic pot, and end with the dynamic `pops' of popcorn expanding, along with the jiggling of a metal around the stove.

Predicting this sound is an instance of \emph{learning from reentry}: where time-locked correlations enable one modality to educate others. Reentry has been hypothesized by developmental psychologists to be crucial for how we as humans learn visual and world knowledge, much of it without need for an explicit teacher \cite{piaget1952origins, edelman1993neural, chapman2000children, smith2005development}. Yet, we ask -- can we build machines that likewise learn vision, language, and sound \emph{together}? And can this paradigm enable learning \emph{neural script knowledge}, that transfers to language-and-vision tasks, \emph{even those without sound}?

In this work, we study these questions, and find that the answers are `\textbf{yes}.' We introduce a new model that learns self-supervised representations of videos, through all their modalities (audio, subtitles, vision). We dub our model \modelnamemedium\footnote{Short for \modelnamelong.}, henceforth \modelname for short. Our model differs from past work that learns from audio-image pairs  \cite{guzhov2021audioclip, lamba2021cross}, from subtitled videos \cite{sun2019videobert, zellers2021merlot}, or from static images with literal descriptions \cite{tan2019lxmert,chen2019uniter,radford2021learning}. Instead, we learn joint representations from \emph{all modalities of a video}, using each modality to teach others. We do this at scale, training on over 20 million YouTube videos.

We introduce a new \emph{contrastive masked span} learning objective to learn script knowledge across modalities. It generalizes and outperforms a variety of previously proposed approaches (e.g. \cite{devlin2018bert, tan2019lxmert, radford2021learning, zellers2021merlot}), while enabling audio to be used as signal. The idea is outlined in Figure~\ref{fig:teaser}: the model must figure out which span of text (or audio) was \masktoken ed out of a video sequence. We combine our objective with a second contrastive learning approach, tailored to learning \emph{visual recognition} from scratch: the model must also match each video frame to a contextualized representation of the video's transcript \cite{zellers2021merlot}. Through ablations, we show that our framework enables rapid pretraining of a model and readily scales to `large' transformer sizes (of 644M parameters).

Experimental results show that \modelname~learns powerful representations, useful even for tasks posed over only a few of the studied modalities.
For example, when finetuned on Visual Commonsense Reasoning \cite{zellers2019recognition} (a vision+language task with no audio), it sets a new state-of-the-art, outperforming models trained on supervised image-caption pairs by \textbf{over 5\%}.
It does even better on video tasks: fine-tuning without audio, it outperforms prior work on TVQA \cite{lei2018tvqa} by a margin of \textbf{over 7\%} (and given TVQA audio, performance increases even further). Finally, audio enables 91.1\% accuracy on Kinetics-600 \cite{carreira2018short}. These performance improvements do not come at the expense of efficiency: our largest model uses one-fifths the FLOPs of a VisualBERT.

\modelname also performs well in zero-shot settings. We evaluate on four diverse benchmarks: Situated Reasoning (STAR) \cite{wu2021star}, EPIC-Kitchens \cite{Damen2021RESCALING}, LSMDC-FiB \cite{lsmdc}, and MSR-VTT QA \cite{xu2017video}. These benchmarks require visual reasoning with respective emphasis on \emph{temporality}, \emph{future prediction}, and both \emph{social} and \emph{physical understanding}. With no fine-tuning or supervision, our model obtains competitive performance on each. Of note, it nearly doubles \cite{yang2020just}'s SoTA zero-shot accuracy on MSR-VTT QA, and it outperforms supervised approaches (like ClipBERT \cite{lei2021less}) on STAR.

Finally, we investigate \emph{why}, and \emph{on which training instances} audio-powered multimodal pretraining particularly helps. For instance, predicting audio rewards models for recognizing \emph{dynamic state changes} (like cooked popcorn) and \emph{human communication dynamics} (what are people's emotions and towards whom). Our model progressively learns these phenomena as pretraining progresses. These signals are often orthogonal to what snippets of text provide, which motivates learning from both modalities.

In summary, our key contributions are the following:
\begin{enumerate}[rowan]
    \item \modelname, a model for multimodal script knowledge, fusing vision, audio, and text.
    \item A new contrastive span matching objective, enabling our model to learn from text \emph{and audio} self-supervision.
    \item Experiments, ablations, and analysis, that demonstrate strong multimodal video representations.
\end{enumerate}
Overall, the results suggest that learning representations from \emph{all modalities} -- in a time-locked, reentrant manner -- is a promising direction, and one that has significant space for future work. We release code and model checkpoints at \websitelink.

\section{Related Work}
\noindent Our work brings together two active lines of research.

\mparagraph{Joint representations of multiple modalities.} Many language-and-vision tasks benefit from \emph{early fusion} of the modalities \cite{baltruvsaitis2018multimodal}. A family of `VisualBERT' models have been proposed for this: typically, these use a supervised object detector image encoder backbone, and pretrain on images paired with literal captions \cite{tan2019lxmert, li2019visualbert, lu2019vilbert, chen2019uniter,  yu2020ernie, lei2021less}. Cross-modal interactions are learned in part through a \emph{masked language modeling} (mask LM) objective \cite{devlin2018bert}, where subwords are replaced with `\masktoken', and models independently predict each subword conditioned on both images and unmasked tokens.\footnote{Recent papers propose extensions, like generating masked-out spans \cite{cho2021unifying} or text \cite{lin2021vx2text, wang2021simvlm}, but it is unclear whether they can outperform the VisualBerts on vision-language tasks like VCR \cite{zellers2019recognition}. Another extension involves learning from text-to-speech audio in a captioning setting \cite{ilharco2019large,liu2021opt} -- yet this lacks key  supervision for environmental sounds and emotive speech.}

Perhaps closest to our work is MERLOT \cite{zellers2021merlot}, which learns a joint vision-text model from web videos with automatic speech recognition (ASR). Through a combination of objectives (including a variant of mask LM), MERLOT established strong results on a variety of video QA benchmarks when finetuned. However, it lacks audio: it is limited to representing (and learning from) video frames paired with subtitles. Our proposed \modelname, which represents and learns from audio, outperforms MERLOT.

\mparagraph{Co-supervision between modalities.} A common pitfall when training a joint multimodal model is that complex \emph{inter-modal} interactions can be ignored during learning, in favor of simpler \emph{intra-modal} interactions \cite{goyal2017making,clark-etal-2019-dont,hessel2020emap}. For example, when using the aforementioned mask LM objective, models can \emph{ignore visual input completely} in favor of text-text interactions \cite{bitton2021data}; this issue is magnified when training on videos with noisy ASR text \cite{zellers2021merlot}.

A line of recent work thus learns independent modality-specific encoders, using objectives that cannot be shortcutted with simple intra-modal patterns. Models like CLIP learn image classification by matching images with their captions, contrastively \cite{zhang2020contrastive, radford2021learning, jia2021scaling}. Recent work has explored this paradigm for matching video frames with their transcripts \cite{xu2021videoclip}, with their audio signal \cite{rouditchenko2020avlnet, wang2021multimodal}, or both \cite{alayrac2020self, akbari2021vatt}; these works likewise perform well on single-modality tasks like audio classification and activity recognition. These independent encoders can be combined through late fusion \cite{rouditchenko2020avlnet}, yet late fusion is strictly less expressive than our proposed joint encoding (early fusion) approach.

\mparagraph{Our work} combines both lines of research. We learn a model for jointly representing videos, through all their modalities, and train it using a new learning objective that enables \emph{co-supervision} between modalities.

\section{Model: \modelname}
\label{sec:modelsection}

In this section, we present \modelname, including: our model architecture (\ref{ssec:modelarch}), new pretraining objectives (\ref{ssec:contrastiveobj}), and pretraining video dataset (\ref{ssec:dset}). At a high level, \modelname~represents a video by fusing its constituent modalities (vision, audio, and text from transcribed speech) together, and over time. These representations enable both finetuned and zero-shot downstream applications.

More formally, we split a video $\mathcal{V}$ into a sequence of non-overlapping segments in time $\{\boldsymbol{s}_t\}$. Each segment has:
\begin{enumerate}[rowan]
    \item A frame $\boldsymbol{v}_t$, from the middle of the segment,
    \item The ASR tokens $\boldsymbol{w}_t$ spoken during the segment,
    \item The audio $\boldsymbol{a}_t$ of the segment.
\end{enumerate}

Segments default to 5 seconds in length; we discuss details of how we split videos into segments in Appendix~\ref{supp:pretrainingdatainfo}. 

As the text $\boldsymbol{w}_t$ was automatically transcribed by a model given audio $\boldsymbol{a}_t$, it is reasonable to assume that it contains strictly less information content.\footnote{Despite being derived from the audio, pretraining with text is still paramount: 1) in \S \ref{ssec:sec_with_reason_for_text_being_included} we discuss how jointly modeling audio+text prevents models from shortcutting pretraining objectives via surface correlations; 2) in \S\ref{sec:sec_with_tvqa_results} we show that incorporating both transcripts and audio during fine-tuning improves performance; and 3) a textual interface to the model is required for downstream vision+language with textual inputs.}
Thus, for each segment $\boldsymbol{s}_t$, we provide models with exactly one of text \emph{or} audio. 
We will further \emph{mask out} portions of the text and audio during pretraining, to challenge models to recover what is missing.


\subsection{Model architecture}
\label{ssec:modelarch}
\begin{figure}[t!]
\vspace{-1mm}\centering\small
 \iftoggle{cvprfinal}{\includegraphics[width=\textwidth]{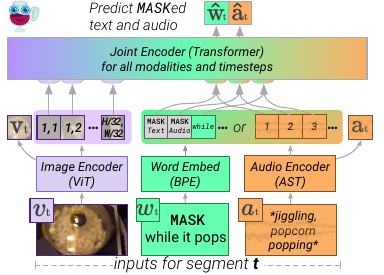}}{\includegraphics[width=\textwidth]{figures/reservefig-modelonly-v2.pdf}}\vspace*{-1mm}
\caption{\modelnamehalfwidth~architecture. We provide sequence-level representations of video frames, and \emph{either} words or audio, to a joint encoder. The joint encoder contextualizes over modalities and segments, to predict what is behind \masktoken~for audio $\widehat{\mathbf{a}_t}$ and text $\widehat{\mathbf{w}_t}$. We supervise these predictions with independently encoded targets: $\mathbf{a}_t$ from the audio encoder, and $\mathbf{w}_t$ from a separate text encoder (not shown).}
  \label{fig:reservemodel}
\end{figure}


An overview of \modelname is shown in Figure~\ref{fig:reservemodel}. We first pre-encode each modality independently (using a Transformer \cite{vaswani2017attention} or images/audio; a BPE embedding table for text). We then learn a joint encoder to fuse all representations, together and over time.

\textbf{Image encoder}. We use a Vision Transformer (ViT; \cite{dosovitskiy2020image}) to encode each frame independently. We use a patch size of 16 and apply a 2x2 query-key-value attention pool after the Transformer, converting an image of size $H{\times}W$ into a $H/32{\times}W/32$ feature map of dimension $d_h$.

\textbf{Audio encoder}. We split the audio in each segment $\boldsymbol{a}_t$ into three equal-sized \emph{subsegments}, for compatibility with the lengths at which we mask text (Appendix~\ref{supp:pretrainingdatainfo}). We use an Audio Spectrogram Transformer to encode each subsegment independently \cite{gong2021ast}. The three feature maps are concatenated; the result is of size $18{\times}d_h$ for every 5 seconds of audio.

\mparagraph{Joint encoder.} Finally, we jointly encode all modalities (over all input video segments) using a bidirectional Transformer. We use a linear projection of the final layer's hidden states for all objectives (e.g. $\widehat{\mathbf{w}}_t$ and $\widehat{\mathbf{a}}_t$).



\mparagraph{Independently-encoded targets.} We will supervise the joint encoder by simultaneously learning independently-encoded `target' representations for each modality. Doing this is straightforward for the image and audio encoders: we add a \clstoken~to their respective inputs, and extract the final hidden state $\mathbf{v}_t$ or $\mathbf{a}_t$ at that position. For text, we learn a separate bidirectional Transformer \emph{span encoder}, which computes targets $\mathbf{w}_t$ from a \clstoken~and embedded tokens of a candidate text span. This enables zero-shot prediction (\ref{ssec:sec_with_zero_shot_results}).

\mparagraph{Architecture sizes.} We consider two model sizes in this work, which we pretrain from random initialization:
\begin{enumerate}[rowan, label=\textbf{\arabic*.}]
\item \basemodel, with a hidden size of 768, a 12-layer ViT-B/16 image encoder, and a 12-layer joint encoder.
\item \largemodel, with a hidden size of 1024, a 24-layer ViT-L/16 image encoder, and a 24-layer joint encoder.
\end{enumerate}
We always use a 12-layer audio encoder, and a 4-layer text span encoder. Details are in Appendix~\ref{supp:modelimpldetails}.

\subsection{Contrastive Span Training}
\label{ssec:contrastiveobj}
\label{ssec:sec_with_reason_for_text_being_included}

\begin{figure}[t!]
  \centering\small\vspace{-1mm}
    \includegraphics[width=\textwidth]{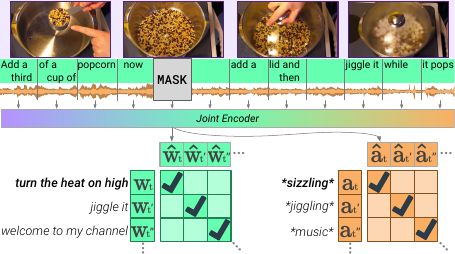}\vspace*{-2mm}%
\caption{Contrastive span training. Given a video with all modalities temporally aligned, we \masktoken~out a region of text and audio. The model must maximize its similarity \emph{only to} an independent encoding of the text $\mathbf{w}_t$ and audio $\mathbf{a}_t$.
}
  \label{fig:reserveobjectives}
\end{figure}

We introduce \emph{contrastive span} training, which enables learning across and between the three modalities. As shown in Figure~\ref{fig:reserveobjectives},
the model is given a sequence of video segments. For each one, we include the video frame, and then three `subsegments' that are each either text \emph{or} audio.
The subdivided audio segments are encoded independently by the Audio Encoder, before being fused by the Joint Encoder.
We train by replacing 25\% of these text and audio subsegments with a special \masktoken~token. The model must match the representation atop the \masktoken~\emph{only with} an independent encoding of its span. 

Our approach combines past success at matching images to their captions \cite{radford2021learning, jia2021scaling} along with `VisualBERT'-style prediction of independent tokens \cite{tan2019lxmert, chen2019uniter} -- though, crucially, we predict representations at a higher-level semantic unit than individual tokens. Our approach also enables the model to learn from both audio and text, while discouraging \emph{memorization} of raw perceptual input, or tokens -- which can harm representation quality \cite{walker2016uncertain}.

Formally, we minimize the cross entropy between the \masktoken ed prediction $\hat{\mathbf{w}}_t$ and its corresponding phrase representation $\mathbf{w}_t$, versus others in the batch $\mathcal{W}$:%
\begin{equation}
\label{eqn:contrastive}
\mathcal{L}_{\textrm{mask}\rightarrow \textrm{text}} {=} \frac{1}{|\mathcal{W}|}\hspace*{-1mm}\sum_{
\mathbf{w}_t \in \mathcal{W}}\hspace*{-1mm}\left(\log \frac{\exp(\sigma \hat{\mathbf{w}}_t \cdot \mathbf{w}_t)}{\sum_{\mathbf{w} \in \mathcal{W}} \exp(\sigma \hat{\mathbf{w}}_t \cdot \mathbf{w})} \right).
\end{equation}%
We first L$^{2}$-normalize $\mathbf{w}$ and $\hat{\mathbf{w}}$, and scale their dot product with a parameter $\sigma$ \cite{radford2021learning}.\footnote{Following past work, we optimize $\sigma$ and clip it at 100, which enables the model to `warm-up' its emphasis placed on hard negatives \cite{radford2021learning, wang2021understanding}.}
We then add this to its transposed version $\mathcal{L}_{\textrm{text}\rightarrow\textrm{mask}}$, giving us our text-based loss $\mathcal{L}_{\textrm{text}}$. Analogously, we define $\mathcal{L}_{\textrm{audio}}$ for audio, between the \masktoken ed prediction $\hat{\mathbf{a}}_t$ and its target $\mathbf{a}_t$, versus others $\mathbf{a}$ in the batch.

In addition to these masked text and audio objectives, we simultaneously train the model to match video frames with a contextualized encoding of the transcript.\footnote{In MERLOT \cite{zellers2021merlot}, this objective was found to be critical for learning visual recognition from self-supervised videos.} Here, the joint encoder encodes the entire video's transcript at once, extracting a single hidden representation per segment $\hat{\mathbf{v}}_t$. We use the same contrastive setup as Equation~\ref{eqn:contrastive} to maximize the similarity of these vectors with the corresponding $\mathbf{v}_t$ vectors from the frames, giving us a symmetric frame-based loss $\mathcal{L}_{\textrm{frame}}$. The final loss is the sum of the component losses:%
\begin{equation}\label{eqn:total}
    \mathcal{L} = \mathcal{L}_{\textrm{text}} + \mathcal{L}_{\textrm{audio}} + \mathcal{L}_{\textrm{frame}}.
\end{equation}%

\mparagraph{Avoiding shortcut learning.} Early on, we observed that training a model to predict a \emph{perceptual} modality (like audio or vision) given input from \emph{the same modality}, led to shortcut learning -- a low training loss, but poor representations. We hypothesize that this setup encourages models to learn imperceptible features, like the exact model of the microphone, or the chromatic aberration of the camera lens \cite{doersch2015unsupervised}. We avoid this, while still using audio as a target, by simultaneously training on two kinds of masked videos:
\begin{enumerate}[rowanroman]
\item\textbf{Audio only as target.} We provide only video frames and subtitles. The model produces representations of both \emph{audio} and \emph{text} that fill in \masktoken ed blanks.
\item\textbf{Audio as input.} We provide the model video frames, and subtitles \emph{or audio} at each segment. Because the model is given audio as an input somewhere, the model only produces representations for \masktoken ed \emph{text}.
\end{enumerate}

Another issue is that YouTube's captions are not perfectly time-aligned with the underlying audio. During our initial exploration, models took ready advantage of this shortcut: for instance, predicting an audio span based on what adjacent (overlapping) words sound like. We introduce a masking algorithm to resolve this; details in Appendix~\ref{supp:pretrainingdatainfo}.

\mparagraph{Pretraining setup.} We train on TPU v3-512 accelerators; training takes 5 days for \basemodel, and 16 days for \largemodel. We made pretraining more efficient through several algorithmic and implementation improvements. Of note, we simultaneously train on written (web) text, which enables more text candidates to be used.
We use a batch size of 1024 videos, each with $N{=}16$ segments (split into two groups of 8 segments each). 
We use AdamW \cite{Kingma2014AdamAM, loshchilov2017decoupled} to minimize Equation~\ref{eqn:total}. More details and hyperparameters are in Appendix~\ref{supp:modelimpldetails}.

\subsection{Pretraining Dataset}
\label{ssec:dset}
Recent prior work on static images that demonstrates empirical improvements by increasing dataset size -- all the way up to JFT-3B \cite{kolesnikov2020big, dosovitskiy2020image, radford2021learning, zhai2021scaling}. The same pattern emerges in videos: prior work that has shown promising empirical improvements not only by scaling to 6 million videos/180M frames \cite{zellers2021merlot}, but also by collecting a diverse set (i.e., going beyond instructional videos \cite{hessel-etal-2019-case}).

To this end, we introduce a new training dataset of 20 million English-subtitled YouTube videos, and 1 billion frames, called \datasetname.
At the same time, we take steps to protect user privacy, directing scraping towards public, large, and monetized channels. We detail our collection, preprocessing, and release strategy in Appendix \ref{supp:datacollection}.

\section{Experiments}
In this section, we present model ablations (\ref{sssec:vcrablations}), and show that a finetuned \modelname~obtains state-of-the-art results on VCR  (\ref{ssec:sec_with_vcr_sota_results}), TVQA (\ref{sec:sec_with_tvqa_results}), and Kinetics-600 (\ref{sec:sec_with_kinetics_results}). We then show that our model has strong zero-shot capability, over four challenging zero-shot tasks (\ref{sec:sec_with_tvqa_results}).

\subsection{Visual Commonsense Reasoning (VCR)}
\label{ssec:vcr}
We evaluate \modelname~first through finetuning on VCR \cite{zellers2019recognition}. Most competitive models for VCR are pretrained exclusively on images paired with captions, often with supervised visual representations (e.g. from an object detector). To the best of our knowledge, the only exception is MERLOT \cite{zellers2021merlot}, which uses YouTube video frames and text as part of pretraining; no VCR model to date was pretrained on audio.

\mparagraph{VCR Task.} A model is given an image from a movie, and a question. The model must choose the correct answer given four multiple choice options ($Q{\rightarrow}A$); it then is given four \emph{rationales} justifying the answer, and it must choose the correct one ($QA{\rightarrow}R$). The results are combined with a $Q{\rightarrow}AR$ metric, where a model must choose the right answer \emph{and then} the right rationale, to get the question `correct.' 

\mparagraph{Finetuning approach.} We follow \cite{zellers2021merlot}'s approach: `drawing on' VCR's detection tags onto the image, and jointly finetuning on $Q{\rightarrow}A$ and $QA{\rightarrow}R$. For both subproblems, we learn by scoring each $Q{\rightarrow}A$ (or $QA{\rightarrow}R$) option independently.
We pool a hidden representation from a \masktoken~inserted after the text, and pass this through a newly-initialized linear layer to extract a logit, which we optimize through cross-entropy (details in Appendix~\ref{supp_ssec:vcrdetails}.)

\vspace{-3mm}
\subsubsection{Ablations: contrastive learning with audio helps.}
\label{sssec:vcrablations}
While we present our final, state-of-the-art VCR performance in \ref{ssec:sec_with_vcr_sota_results}, we first use the corpus for an ablation study.
We use the same architecture and data throughout, allowing apples-to-apples comparison between modeling decisions. We start with a similar configuration to MERLOT \cite{zellers2021merlot} and show that contrastive span training improves further, particularly when we add audio.

\mparagraph{Contrastive Span helps for \emph{V}ision+\emph{T}ext modeling.} 
We start by comparing pretraining objectives for learning from YouTube ASR and video alone:
\begin{enumerate}[rowan]
\item \textbf{Mask LM.} This objective trains a bidirectional model by having it \emph{independently} predict masked-out tokens. We make this baseline as strong as possible by using SpanBERT-style masking \cite{joshi2020spanbert}, where text spans are masked out (identical to our \emph{contrastive spans}). Each span $\boldsymbol{w}$ is replaced by a \masktoken~token, and we predict each of its subwords $w_i$ independently.\footnote{Like \cite{joshi2020spanbert}, we concatenate the \masktoken's hidden state with a position embedding for index $i$, pass the result through a two-layer MLP, and use tied embedding weights to predict $w_i$.}
\item \textbf{VirTex} \cite{desai2021virtex}. In this objective, we likewise mask text subsegments and extract their hidden states. The difference is that we sequentially predict tokens $w_i \in \boldsymbol{w}$, using a left-to-right language model (LM) with the same architecture details as our proposed span encoder. 
\end{enumerate}
\begin{table}[t!]
{\footnotesize
\vspace{-2mm}
\resizebox{.99\textwidth}{!}{\begin{tabular}{@{} p{0.3cm} @{\hspace{0.05cm}} p{5.5cm} @{\hspace{0.05cm}} p{0.95cm} @{\hspace{0.01cm}} p{0.5cm} @{}}
& Configuration \newline {\footnotesize \emph{for one epoch of pretraining}} & VCR Q$\rightarrow$A & {\footnotesize val} {\footnotesize (\%)}\\ \cmidrule{1-3}
\multirow{3}{*}{\rotatebox[origin=c]{90}{{\footnotesize V+T}}}
& Mask LM \cite{devlin2018bert, tan2019lxmert, zellers2021merlot} & 67.2 \\
& VirTex-style \cite{desai2021virtex} & 67.8 \\
& \wineglassemoji~Contrastive Span & \textbf{69.7} \\ \cmidrule{1-3}
\multirow{4}{*}{\rotatebox[origin=c]{90}{{\footnotesize V+T+A}}}
& \wineglassemoji~Audio as target & 70.4 \\
& \wineglassemoji~Audio as input and target & \textbf{70.7} \\
& Audio as input and target, w/o strict localization & 70.6 \\ \cmidrule{2-3}
& \basemodel & \textbf{71.9} \\
\end{tabular} 
}
\caption{\textbf{Ablation study} of our contrastive span objective. It outperforms prior work in a Vision+Text setting, with a 1\% boost when audio is added. Our full setup, adding written text, improves another 1\%. \wineglassemoji~denotes part of our full model. \label{tab:ablations}}}\vspace{-1mm}
\end{table}

Results are in Table~\ref{tab:ablations}. Versus these approaches, our contrastive span objective boosts performance by over 2\%, after one epoch of pretraining only on vision and text. We hypothesize that its faster learning is caused by encouraging models to learn concept-level span representations; this might not happen when predicting tokens individually \cite{cho2021tweet}. 

\mparagraph{\emph{A}udio pretraining helps}, even for the audio-less VCR:
\begin{enumerate}[rowan]
\setcounter{enumi}{3}
\item \textbf{Audio as target.} Here, the model is only given video frames and ASR text as input. In addition to performing contrastive-span pretraining over the missing text spans, it does the same over the (held-out) audio span (Equation~\ref{eqn:total}. This boosts VCR accuracy by 0.7\%.
\item \textbf{Audio as input and target.} The model does the above (for video+text input sequences), and simultaneously is given video+text+audio sequences, wherein it must predict missing text. This boosts accuracy by 1\% in total.
\item \textbf{Sans strict localization.} We evaluate the importance of our strict localization in time. Here, in addition to correct subsegments at the \emph{true} position $t$ as a correct match, we count adjacent \masktoken ed out regions as well. An extreme version of this was proposed by \cite{gordon2020watching}, where a positive match can be of any two frames in a video. Yet even in our conservative implementation, performance drops slightly, suggesting localization helps.
\end{enumerate}
Putting these all together, we find that contrastive span pretraining outperforms mask LM, with improved performance when audio is used \textbf{both as input and target}. For our flagship model, we report results in Table~\ref{tab:ablations} on simultaneously training on web-text sequences as well (Appendix~\ref{supp_ssec:puttingitalltogether}), this improves performance by an additional 1\%.

\begin{figure*}[t!]
\vspace{-2mm}
\begin{floatrow}
\captionsetup{font=footnotesize}

\hspace{-3mm}\ffigbox[0.3\textwidth]{\caption{\textbf{Pretraining progress}: performance on contrastive-span pretraining, vs. finetuned VCR validation accuracy. Pretraining \basemodel for 9 more epochs boosts performance by 5\%; \maskblockyfont{L} by 8\%.
}\label{fig:pretrainingprogress}}{\includegraphics[width=0.9\linewidth]{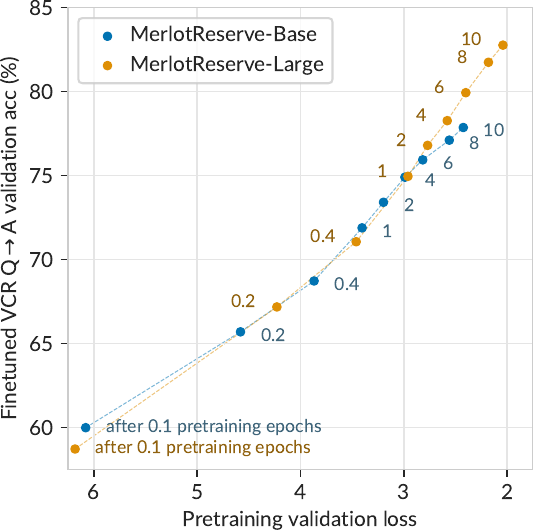}\vspace{-2mm}}%
\hspace{-2mm}
\capbtabbox{\caption{\modelname gets \textbf{state-of-the-art leaderboard performance on VCR}. We compare it with the largest submitted single models, including image-caption models that utilize heavy manual supervision (e.g. object detections and captions).}\label{tab:vcrsota}}{%
\resizebox{.35\textwidth}{!}{
\begin{tabular}{@{} p{0.35cm} @{\hspace{0.05cm}} p{4cm} @{\hspace{0.02cm}} p{1cm} @{\hspace{0.02cm}} p{1cm} @{\hspace{0.02cm}} p{0.8cm}@{}}
&              & \multicolumn{3}{c}{\scriptsize VCR test (acc; \%)} \\ 
& Model & {\scriptsize Q$\rightarrow$A} & {\scriptsize QA$\rightarrow$R} & {\scriptsize Q$\rightarrow$AR} \\ \toprule
\multirow{7}{*}{\rotatebox[origin=c]{90}{{\footnotesize Caption/ObjDet-based}}} 
& ERNIE-ViL-Large \cite{yu2020ernie} & 79.2 & 83.5 & 66.3 \\
& Villa-Large \cite{gan2020large} & 78.9 & 83.8 & 65.7 \\
& UNITER-Large \cite{chen2019uniter}& 77.3 & 80.8 & 62.8 \\
& Villa-Base \cite{gan2020large}& 76.4 & 79.1 & 60.6 \\
& VilBERT \cite{lu2019vilbert} & 73.3 & 74.6 & 54.8 \\
& B2T2 \cite{alberti2019fusion} & 72.6 & 75.7 & 55.0 \\ 
& VisualBERT \cite{li2019visualbert} & 71.6 & 73.2 & 52.4 \\ \midrule
\multirow{3}{*}{\rotatebox[origin=c]{90}{{\footnotesize Video-based}}} & MERLOT \cite{zellers2021merlot} & 80.6 & 80.4 & 65.1 \\ \cmidrule{2-5}
& \basemodel & 79.3 & 78.7 & 62.6 \\ 
& \largemodel & \textbf{84.0} & \textbf{84.9} & \textbf{72.0} \\  \midrule
\end{tabular}\vspace{-3mm}
}}
\hspace{5mm}
\capbtabbox{\caption{ \modelname~gets state-of-the-art results on TVQA by \textbf{over 7\%}, versus prior work (that cannot make use of audio).\label{tab:tvqatable} }} %
{%
\resizebox{.27\textwidth}{!}{\begin{tabular}{@{} p{0.4cm} @{\hspace{0.05cm}} p{3.4cm} @{\hspace{0.05cm}} p{0.95cm} @{\hspace{0.01cm}} p{0.95cm} @{}}
&       &\multicolumn{2}{c}{\hspace{-2mm}{\scriptsize TVQA (acc; \%) }} \\
& Model & Val & Test \\ \toprule
& Human \cite{lei2018tvqa} & -- & 89.4 \\ \midrule
\multirow{5}{*}{\rotatebox[origin=c]{90}{{\footnotesize Subtitles}}}  & MERLOT \cite{zellers2021merlot} & 78.7 & 78.4 \\
& MMFT-BERT \cite{urooj2020mmft} & 73.5 & 72.8 \\
& Kim et al \cite{Kim2020SelfsupervisedPA} & 76.2 & 76.1 \\
& \basemodel         & 82.5 & -- \\
& \largemodel         & \textbf{85.9}    & \textbf{85.6} \\ \midrule
\multirow{2}{*}{\rotatebox[origin=c]{90}{{\footnotesize Audio}}} 
& \basemodel     & 81.3 & --      \\
& \largemodel     & \textbf{85.6}     & \textbf{84.8}  \\ \midrule
\multirow{2}{*}{\rotatebox[origin=c]{90}{{\footnotesize Both}}} 
& \basemodel      & 83.1        &  82.7\\
& \largemodel    & \textbf{86.5} & \textbf{86.1} \\ \midrule
\end{tabular}\vspace{-3mm}}}%

\end{floatrow}
\end{figure*}

\vspace{-4mm}
\subsubsection{VCR Results}\vspace{-2mm}
\label{ssec:sec_with_vcr_sota_results}
Encouraged by these results, we train our models for 10 epochs on \datasetname. Figure~\ref{fig:pretrainingprogress} demonstrates that finetuned VCR performance tracks with the number of pretraining epochs, as well as the validation loss.\footnote{The plot suggests that if we pretrained longer, VCR performance might continue to increase, though a confounding factor might be the learning-rate schedule. With access to compute beyond our current capacity, future work would be well-suited to consider this and other pre-training modifications.
}

Finally, in Table~\ref{tab:vcrsota}, we compare \modelname~against the largest published models from the VCR leaderboard. Of note, \largemodel outperforms all prior work, by \textbf{over 5\%} on Q$\rightarrow$AR metric. It outperforms even large ensembles (e.g. 15 ERNIE-Large's) submitted by industry \cite{yu2020ernie}, though we do not show these on this table to focus on only single models. 

\textbf{Efficiency.} The accuracy increase of \modelname~is not simply due to compute.\footnote{Here, we use FLOPs as our key efficiency metric, as they are a critical bottleneck in model scaling \cite{kaplan2020scaling, dosovitskiy2020image, zhai2021scaling}. On the other hand, we argue that parameter count can be misleading -- for instance, many Transformer parameters can be tied together with minimal performance loss \cite{lan2019albert}.}
In fact, our \largemodel requires \emph{one-fifth the FLOPs} of detector-based systems, like UNITER-Large \cite{chen2019uniter} (Appendix~\ref{supp_ssec:efficiency}).
Moreover, because \largemodel~uses a pure ViT backbone versus MERLOT's ViT-ResNet hybrid, it uses fewer FLOPs than MERLOT, while scoring 7\% higher. Meanwhile, \basemodel~outperforms `base' detector-based models, while using \emph{less than one-tenth their FLOPs}. 

In terms of parameter count, \basemodel is comparable to prior work. On VCR, including the vision stack, \basemodel~has 200M finetunable parameters and performs similarly to the 378M parameter UNITER-Large. \largemodel~has 644M parameters.

\subsection{Finetuning on TVQA}
\label{sec:sec_with_tvqa_results}
Next, we use TVQA \cite{lei2018tvqa} to evaluate our model's capacity to transfer to multimodal video understanding tasks. In TVQA, models are given a video, a question, and five answer choices. The scenes come from American TV shows, and depict characters interacting with each other through dialogue -- which past work represents through subtitles. 

\mparagraph{Audio-Subtitle Finetuning.} To evaluate how much audio can help for TVQA, we finetune \modelname~jointly between the `Subtitles' and `Audio' settings. Like on VCR, we consider one sequence per candidate: each contains video frame features, the question, the answer candidate, and a \masktoken~token (from where we pool a hidden representation). During training, each sequence is duplicated: we provide one sequence with \emph{subtitles} from the video, and for the other, we use \emph{audio}. This lets us train a single model, and then test how it will do \emph{given subtitles}, \emph{given audio}, or \emph{given both} (by averaging the two softmax predictions).

\mparagraph{Results.} We show TVQA results in Table~\ref{tab:tvqatable}. With subtitles and video frames alone, our \basemodel outperforms all prior work by over 3\%. Combining subtitle-only and audio-only predictions performs even better, improving over 4\% versus the prior state-of-the-art, MERLOT (and in turn over other models). The same pattern holds (with additional performance gains) as model size increases: \largemodel~improves over prior work by \textbf{7.6\%}.

\newcommand{\nodata}[1]{\multicolumn{#1}{c}{\cellcolor{gray!10}}}
\newcommand{\nodataright}[1]{\multicolumn{#1}{|c}{\cellcolor{gray!10}}}
\newcommand{\nd}{\cellcolor{gray!10}}
\newcommand\todo[1]{\cellcolor{red!10} #1}

\begin{figure*}[t!]
\vspace*{-3mm}\begin{floatrow}
\captionsetup{font=footnotesize}

\capbtabbox{\caption{\modelname~gets state-of-the-art results on Kinetics-600 by \textbf{1.5\%} versus standard approaches (that cannot make use of audio).\label{tab:kinetics} }} %
{%
\resizebox{.22\textwidth}{!}{
  \begin{tabular}{@{} p{0.4cm} @{\hspace{0.05cm}} p{3.4cm} @{\hspace{0.05cm}} p{0.95cm} @{\hspace{0.01cm}} p{0.9cm} @{}}
\multicolumn{4}{r}{\hspace{2mm}{\scriptsize Kinetics-600 (\%) }} \\
&  Model & Top-1 & Top-5 \\ \toprule
\multirow{9}{*}{\rotatebox[origin=c]{90}{{\footnotesize Vision Only}}} 
& VATT-Base\cite{akbari2021vatt} & 80.5 & 95.5 \\
& VATT-Large \cite{akbari2021vatt} & 83.6 & 96.6 \\
& TimeSFormer-L \cite{bertasius2021space} & 82.2 & 95.6 \\
& Florence \cite{yuan2021florence} & 87.8 & 97.8 \\
& MTV-Base \cite{yan2022multiview} & 83.6 & 96.1 \\
& MTV-Large \cite{yan2022multiview} & 85.4 & 96.7 \\
& MTV-Huge \cite{yan2022multiview} & 89.6 & 98.3 \\
& \basemodel         & 88.1 & 95.8 \\
& \largemodel         & 89.4  & 96.3  \\ \midrule
\multirow{2}{*}{\rotatebox[origin=c]{90}{{\footnotesize +Audio}}} 
& \basemodel          & 89.7 & 96.6 \\
& \largemodel          & \textbf{91.1} & \textbf{97.1}  \\ \midrule
\end{tabular}\vspace{-2mm}}}
\hspace{-2mm}
\capbtabbox{ \caption{Zero shot results. On STAR, \modelname obtains state-of-the-art results, outperforming finetuned video models. It performs well on EPIC-Kitchens (verb and noun forecasting), along with LSMDC, despite their long-tail distributions. On MSR-VTT QA, it outperforms past work on weakly-supervised video QA. Further, it outperforms CLIP (that cannot handle dynamic situations), and benefits from audio when given. \label{tab:zsresults} }} %
{%
\newcolumntype{P}{>{\centering\arraybackslash}p{0.6cm}}
\resizebox{.70\textwidth}{!}{
\begin{tabular}{@{} p{0.35cm} @{} p{4.0cm} @{\hspace{0.05cm}} P@{\hspace{1em}}P@{\hspace{1em}}P@{\hspace{1em}}P@{\hspace{1em}}P@{\hspace{4em}}
                                                              P@{\hspace{1em}}P@{\hspace{1em}}P@{\hspace{4em}}
                                                              P@{\hspace{4em}} 
                                                              P@{\hspace{1em}}P}
    & & \multicolumn{5}{l}{\parbox{4.2cm}{\centering Situated Reasoning (STAR)\newline {\scriptsize (test acc; \%)}}} 
        & \multicolumn{3}{l}{\hspace{-1.25em}\parbox{3cm}{\centering EPIC-Kitchens \newline {\scriptsize (val class-mean R\@5; \%)}}} 
        & \multicolumn{1}{l}{\hspace{-1.6em}\parbox{1.3cm}{LSMDC \newline {\scriptsize (FiB test \%)}}}  
        & \multicolumn{2}{l}{\hspace{-1.6em}\parbox{2.3cm}{\centering MSR-VTT QA \newline {\scriptsize (test acc \%)}}}   \\
\multicolumn{2} {c}{Model} & {\centering\tiny Interaction} & {\centering\tiny Sequence} & {\centering\tiny Prediction} & {\centering\tiny Feasibility} & {\centering\scriptsize Overall} &  {\centering\scriptsize Verb} & {\centering\scriptsize Noun} & {\centering\scriptsize Action} & {\centering\scriptsize Acc} & {\centering\scriptsize top1} & {\centering\scriptsize top5} \\ \toprule
& \multirow{2}{*}{Supervised SoTA}                & \multicolumn{5}{c}{{\hspace{-2em}\smaller ClipBERT \cite{lei2021less}}} & \multicolumn{3}{c}{{\hspace{-2em}\smaller AVT+ \cite{girdhar2021anticipative}}} & \multicolumn{3}{c}{{\hspace{-2em}\smaller  MERLOT \cite{zellers2021merlot}}} \\
&  & 39.8 & 43.6 & 32.3 & 31.4 & 36.7 & 28.2 & 32.0 & 15.9 & 52.9 & 43.1 & \nodata{1} \\ 
\midrule
\multirow{7}{*}{\rotatebox[origin=c]{90}{{\footnotesize zero-shot}}} 
& Random & 25.0 & 25.0 & 25.0 & 25.0 & 25.0 & \phantom{0}6.2 & \phantom{0}2.3 & \phantom{0}0.1 & \phantom{0}0.1 & \phantom{0}0.1 & \phantom{0}0.5 \\
& CLIP (VIT-B/16) \cite{radford2021learning} & 39.8 & 40.5  & 35.5  & 36.0 & 38.0  & 16.5 & 12.8  & \phantom{0}2.3  & \phantom{0}2.0  & \phantom{0}3.0  & 11.9 \\
& CLIP (RN50x16) \cite{radford2021learning} & 39.9 & 41.7  & 36.5  & \textbf{37.0} & 38.7 & 13.4 & 14.5  & \phantom{0}2.1  & \phantom{0}2.3 & \phantom{0}2.3  & \phantom{0}9.7 \\
& Just Ask {\smaller(ZS)}\cite{yang2020just} & \nodata{5} & \nodata{3} & \nodata{1} & \phantom{0}2.9 & \phantom{0}8.8 \\ \cmidrule{2-13}
& \basemodel & 44.4 & 40.1  & 38.1  & 35.0  & 39.4 & 17.9 & 15.6 & \phantom{0}2.7 & 26.1  & \phantom{0}3.7  & 10.8 \\
& \largemodel & 42.6 & 41.1  & 37.4  & 32.2  & 38.3 & 15.6 & 19.3 & \phantom{0}4.5 & 26.7  & \phantom{0}4.4  & 11.5 \\
& \basemodel (+audio) & \textbf{44.8} & \textbf{42.4}  & \textbf{38.8}  & 36.2  & \textbf{40.5} & 20.9 & 17.5 & \phantom{0}3.7  & 29.1  & \phantom{0}4.0  & 12.0 \\
& \largemodel (+audio) & 43.9 & 42.6  & 37.6  & 33.6  & 39.4 & \textbf{23.2} & \textbf{23.7} & \phantom{0}\textbf{4.8}  & \textbf{31.0}  &  \phantom{0}\textbf{5.8}  & \textbf{13.6} \\
\midrule
\end{tabular}\vspace{-2mm}%
}}
\end{floatrow}

\end{figure*}

\subsection{Finetuning on Kinetics-600 Activity Recognition}
\label{sec:sec_with_kinetics_results}

Next, we use Kinetics-600 \cite{carreira2018short} to compare our model's (finetuned) activity understanding versus prior work, including many top-scoring models that do not integrate audio. The task is to classify a 10-second video clip as one of 600 categories. We finetune \modelname jointly over two settings: vision only, and vision+audio.

\mparagraph{Results.} We show Kinetics-600 results on the validation set, in Table~\ref{tab:kinetics}. \modelname~improves by \textbf{1.7\%} when it can jointly represent the video's frames with its sound. This enables it to outperform other large models, including VATT \cite{akbari2021vatt} which learns to represent audio independently from vision (and so cannot early-fuse them), along with the larger MTV-Huge model \cite{yan2022multiview} by 1.5\%.

\subsection{Zero-Shot Experiments}
\label{ssec:sec_with_zero_shot_results}

Next, we show that our model exhibits strong zero-shot performance for a variety of downstream tasks. Our zero-shot interface is enabled by our \emph{contrastive span objective}. For QA tasks that require predicting an option from a label space of short phrases, we encode this label space as vectors, and predict the closest phrase to a \masktoken ed input. We consider:

\begin{enumerate}[rowanroman]
\item Situated Reasoning (STAR) \cite{wu2021star}. This task requires the model to reason over short situations in videos, covering four axes: interaction, sequence, prediction, and feasibility. The model is given a video, a templated question, and 4 answer choices. We convert templated questions into literal statements (which are more similar to YouTube dialogue); the label space is the set of four options.
\item Action Anticipation in Epic Kitchens \cite{Damen2021RESCALING}. Here, the goal is to predict \emph{future actions} given a video clip, which requires reasoning temporally over an actor's motivations and intentions. The dataset has a long tail of rare action combinations, making zero-shot inference challenging (since we do not assume access to this prior). As such, prior work \cite{girdhar2021anticipative, furnari2020rulstm} trains on the provided in-domain training set. To adapt \modelname~to this task, we provide it a single \masktoken~token as text input, and use as our label space of all combinations of verbs and nouns in the vocabulary (e.g. `cook apple, cook avocado', etc.).
\item LSMDC \cite{maharaj2017dataset,lsmdc}. Models are given a video clip, along with a video description (with a \masktoken~to be filled in). We compare it with the vocabulary used in prior work \cite{zellers2021merlot}.
\item MSR-VTT QA \cite{xu2017video}. This is an open-ended video QA task about what is literally happening in a web video. We use GPT3 \cite{brown2020language}, prompted with a dozen (unlabelled) questions, to reword the questions into statements with \masktoken s. This introduces some errors, but minimizes domain shift. We use a label space of the top 1k options.

\end{enumerate}
For these tasks, we use $N{=}8$ video segments (dilating time when appropriate), and provide audio input when possible. Details and prompts are in Appendix~\ref{supp:downstream}. We compare against both finetuned and zeroshot models, including running CLIP \cite{radford2021learning} on all tasks. CLIP is a strong model for zero-shot classification, particularly when \emph{encyclopedic knowledge about images} is helpful; our comparisons showcase where multimodal script knowledge helps.

\begin{figure*}[t!]
  \centering\small\vspace{-3.4mm}
    \includegraphics[width=\textwidth]{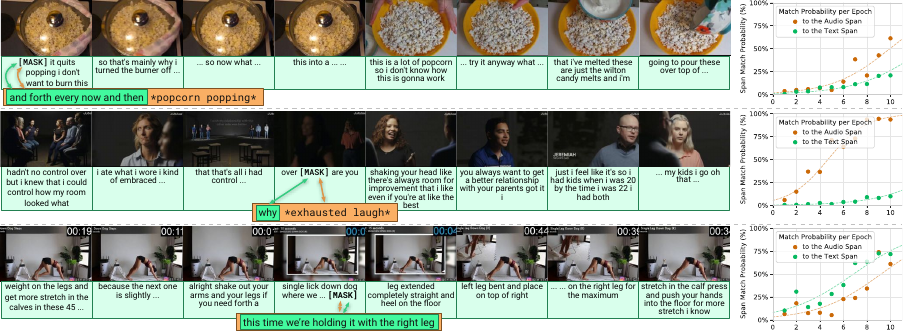}\vspace{-1mm}
\caption{\textbf{Exploring \masktoken ed audio self-supervision}. Shown are example videos from our validation set, with predictions from \basemodel. During pretraining, our model progressively learns to pick up on audio-specific clues. It seems to recognize physical dynamics of \emph{cooking popcorn}, matching the first row to its \masktoken ed audio. Likewise, it seems to use social reasoning to match the second row to its audio. Both of these clues are orthogonal to what the subtitles provide.
}\vspace*{-0.4mm}
  \label{fig:qualfig}
\end{figure*}
\mparagraph{Results.} Table~\ref{tab:zsresults} shows our model performs competitively:
\begin{enumerate}[rowanroman]
\item On STAR, it obtains state-of-the-art results, with performance gain when audio is included. Interestingly, \basemodel~outperforms its larger variant; we hypothesize that this is due to limited prompt searching around question templates. We qualitatively observed that \largemodel~sometimes excludes topically correct options if they sound grammatically strange (to it).
\item On EPIC-Kitchens, our model obtains strong results at correctly anticipating the verb and noun - despite the heavy-tailed nature of both distributions. It is worse on getting both right (`action'), we suspect that this might be due to priors (motifs) between noun and verb \cite{zellers2018scenegraphs}. These are easy to learn given access to training data, but we exclude these as we consider the zero-shot task. 
\item On LSMDC, our model obtains strong results at filling-in-the-blank, likewise despite a heavy (unseen) frequency bias. Notably, it outperforms CLIP significantly, with CLIP often preferring templates that use visually-relevant words, even if they don't make sense as a whole. For instance, given a clip of a mailman, CLIP chooses `the mailman smiles off,' versus `the mailman takes off.'
\item Finally, our model performs well on MSR-VTT QA, outperforming past work that directly rewords subtitled instructional videos into video QA instances \cite{yang2020just}.  
\end{enumerate}

\vspace{-1mm}
\section{Qualitative Analysis: Why does audio help?}
\vspace{-1mm}
What can \modelname learn from both text \emph{and} audio? Three validation set examples are shown in Figure~\ref{fig:qualfig}. The model is given the displayed text and video frames, and must match the \masktoken~to the correct missing text and audio span (out of 48k total in the batch). The plots show \basemodel's probability of correctly identifying the correct audio or text span, as it progresses through 10 epochs of pretraining.

\mparagraph{Audio's supervisory signal.} In the first two rows of Figure~\ref{fig:qualfig}, audio provides orthogonal supervision to text:
\begin{enumerate}[rowan, label=\textbf{\arabic*.}]
\item In the first row, the \masktoken ed audio contains the sound of popcorn pops slowing. By the final epoch, \basemodel~selects this specific auditory cue with 60\% probability, over others (including from adjacent segments, at different stages of popping). Here, sound provides signal for joint vision-text understanding of the situation, as evidenced by its greater match probability. 
\item The second row contains only the text `why,' with the audio providing greatly more information --- a female-presenting speaker (shown in the next frame) laughs, astonished that the child (in the frame afterwards) might want a better relationship with their parents. 
\item In the third row, matching performance is similar between modalities, possibly as the yogi is narrating over a (muted) video recording, and not adding much information.
\end{enumerate}

\mparagraph{Role of text.} Text is still a crucial complement to audio, in terms of the supervision it provides. Consider the second row: \basemodel~learns to match the audio almost perfectly (perhaps reasoning that the speaker is shown in the next frame, and is laughing). In later epochs, its text-match probability increases: knowing that a `why' question is likely to be asked is a valid \emph{social} inference to make about this (tense) situation.

\mparagraph{Learning through multimodal reentry.} Developmental psychologists have hypothesized that human children learn by \emph{reentry}: learning connections between all senses as they interact with the world \cite{edelman1993neural, smith2005development}. Using a held-out modality (like audio) might support learning a better world representation (from e.g. vision and text), by forcing models to abstract away from raw perceptual input. Our work suggests that reentry has potential for machines as well.

%


\section{Conclusion, Limitations, Broader Impact}
We introduced \modelname, which learns jointly through sound, language, and vision, guided through a new pretraining objective. Our model performs well in both finetuned and zero-shot settings, yet it has limitations. Our model only learns from 40-second long videos; relies on ASR models for subtitles, and can only match (not generate) text and audio.  

Still, we foresee broad possible societal impact of this line of work. Video-pretrained models might someday assist low vision or d/Deaf users \cite{leo2017computer, goodman2021toward}. Yet, the same technology can have impacts that we authors consider to be negative, including surveillance, or applications that hegemonize social biases. We discuss these further in Appendix~\ref{supp:broaderimpactstatement}: key dimensions include respecting user privacy during dataset collection, exploring biases in YouTube data, dual use, and energy consumption. We discuss our plan to \emph{release our model and data} for research use so others can critically study this approach to learning script knowledge.%

\section*{Acknowledgements}
{\smaller\begin{spacing}{0.5}
We thank the anonymous reviewers, as well as Jae Sung Park, Oren Etzioni, Gabriel Ilharco, and Mitchell Wortsman for feedback on this work. Thanks also to Zak Stone and the Google Cloud TPU team for providing access to the TPU machines used for conducting experiments. Thanks to James Bradbury and Skye Wanderman-Milne for help with JAX on TPUs. Thanks to the AI2 ReVIZ team, including Jon Borchardt and M Kusold, for help with the demo. This work was funded by DARPA MCS program through NIWC Pacific (N66001-19-2-4031), and the Allen Institute for AI. Last, but not least, thanks to the YouTubers whose work and creativity helps machines to learn about the multimodal world.\end{spacing}}

{\small
\bibliographystyle{ieee_fullname}
\bibliography{main}
}

\appendix
\clearpage

\begin{abstract}
We provide the following materials in the appendix:
\begin{itemize}[labelwidth=!,itemsep=0pt,topsep=1pt,parsep=1pt]
    \item A full broader impact statement (Section~\ref{supp:broaderimpactstatement})
    \item Details about our model architecture (Section~\ref{supp:modelimpldetails})
    \item Details about how we provide video data into the model, including how we align the modalities and perform the masking (Section~\ref{supp:pretrainingdatainfo}
    \item Details about how we adapted our model to downstream tasks (Section~\ref{supp:downstream})
    \item Details about how we collected data (Section~\ref{supp:datacollection})
    \item Additional experiments (Section~\ref{supp:addlexperiments})
\end{itemize}
\end{abstract}

\section{Broader Impact Statement}
\label{supp:broaderimpactstatement}
In this paper, we have presented a model for learning multimodal neural script knowledge, through incorporation of audio as a first-class citizen alongside text and video frames. We argue that academic study of this learning paradigm is important, in part because it relates to how we as humans understand the world. We as humans process situations by perceiving through multiple modalities and interpreting the result holistically. 

At the same time, the work and methodology that we outlined risks dual use. Like other large machine learning systems pretrained on web data, our system may reproduce harmful social biases present in its training data. While a variety of past work has studied risks of \emph{language-only} pretraining \cite{zellers2019grover, bommasani2021opportunities, bender2021dangers, hovy2021five}, the video-centric pretraining that we explore in our work might have different benefits and risks. We discuss these below, along with how we worked to mitigate them through our work.

\subsection{Privacy.} A significant risk with training on data at YouTube scale is protecting user privacy. We took several proactive steps to ensure this, that in turn build off prior work and community norms \cite{abu2016youtube, miech19howto100m, zellers2021merlot}:

\begin{enumerate}[rowan]
\item We release only the video IDs for download, following prior work \cite{abu2016youtube, miech19howto100m}. Thus, if a user deletes a video off of YouTube, it becomes removed from \datasetname~as well, giving content creators a right to opt out of all uses of their videos.
\item Building off of past work \cite{zellers2021merlot}, we directed our data collection towards \emph{public} and \emph{monetized channels}. These channels are identifiable insofar as they contain more subscribers, and more videos. They include companies that have official accounts, including journalism outlets like the \emph{New York Times} and \emph{Vox}. They also include individuals for whom making public YouTube videos is their full time job. In either case, our use videos in question for research purposes can be seen as \emph{fair use}.
\end{enumerate}

\textbf{Framing of privacy.} Privacy is a nuanced topic with many societally, culturally, and generationally-specific interpretations. We took inspiration from Marwick and Boyd \cite{marwick2014networked}'s framework of \emph{networked privacy}, which posits that users posting public videos might \emph{encode} private information -- enough so that their intended viewership (friends, possibly) can catch the gist, but not enough so as to leak private details like phone numbers to the world. 

Through the lens of networked privacy, we see key differences between studying videos on a moderated platform, versus NLP work that trains models from the open web (e.g. \cite{devlin2018bert, raffel2020t5, brown2020language}).
When YouTube users upload videos, they tend to understand details of its privacy policy, beyond consenting to it \cite{kang2015my}. Likewise, YouTubers typically upload their own videos \cite{strangelove2020watching}; the platform deters users from re-posting other users' content. These factors differ from text on the open web. Today, `data brokers' post private details (like phone numbers) to the web for profit \cite{crain2018limits}; concerningly, a study on language models suggests that models are vulnerable at \emph{memorizing} this private information \cite{carlini2020extracting}.

It is worth examining our research through other framings of privacy as well. For example, internet platforms profit off of user data, whereas users do not share equally in these profits \cite{fuchs2011alternative}. For this, and for the other reasons mentioned, we aim to release our model only for research-based use.

\subsubsection{Empirical study: can \modelname identify individual celebrities?}
\label{supp_sssec:celeb}

\begin{table}[t]
\begin{tabular}{@{}lccc@{}}
& \multicolumn{3}{c}{Accuracy (\%)} \\ 
 Model  & Voice & Image+Voice & Image \\ \midrule
 \largemodelhalfwidth & 10.8 & 9.6 & 10.7 \\
 CLIP ViT-B/16 \cite{radford2021learning} & & & \textbf{86.0} \\
\end{tabular}
\caption{Zero-shot person (face/voice) recognition accuracy on VoxCeleb2 \cite{nagrani2017voxceleb} and VGGFace2 \cite{cao2018vggface2}, using different modalities. While \modelnamehalfwidth~can perform person recognition from several modalities, its performance is much lower than the recognition-optimized CLIP model in the image-to-name setting. We hypothesize that this is due to a similarity between this setting and CLIP's pretraining data -- news articles often include celebrity images, paired with their names. }
\label{tab:voxcelebtable}
\end{table}

Inspired by work studying language model memorization of private information \cite{carlini2020extracting}, we wish to empirically probe \modelname's ability to recognize individuals. Our goal during model development was \textbf{not} to optimize for this ability. Instead, our goal was to study models for \emph{multimodal script knowledge} (what people might be doing in a situation over time, and why) instead of long-tailed \emph{visual recognition} (including who those individuals are). These goals might trade off -- for instance, our training data only has individuals' names when they are mentioned in ASR subtitles, a pairing that might be significantly noisier than images and text on the open web.

We study this capacity on the VoxCeleb2 and VGGFace2 datasets \cite{nagrani2017voxceleb, cao2018vggface2}, where we created a test set of 120 celebrities, with 100 samples of each. We study these datasets not to promote them, but to establish a \textbf{conservative upper-bound} for the capacity of a model to recognize non-celebrities. We hypothesize that if \modelname~struggles to select the right celebrity out of 120 predefined options, it would struggle much more at identifying random people (where the set of candidate names is much greater). We test this hypothesis over three zero-shot settings:

\begin{enumerate}[rowan, label=\textbf{\arabic*.}]
\item \textbf{Voice to name.} Given an audio clip sampled for a celebrity, we encode it with our model's audio encoder. We provide our model's joint encoder the text `{\tt\smaller the sound of} \masktoken', followed by the encoded audio. A blank image is provided. We extract the representation on top of the \masktoken, and choose the most similar celebrity name.
\item \textbf{Image+voice to name.} Here, we adopt the same format as `Audio to name,' except we additionally encode an image of the celebrity's face in question.
\item \textbf{Image to name.} Here, \modelname~encodes an image of the celebrity in question, and we provide it with text `A picture of \masktoken.' No audio is provided. Using our model's joint encoder, we select the closest encoded celebrity name, out of all options.

We use this format to compare to a CLIP model, which was trained on web images with captions \cite{radford2021learning}. For the CLIP comparison, we use it to encode each image, and for all considered celebrity names, the sentence `A picture of {\tt \$\{name\}}'. We choose the closest encoded sentence to the encoded image.
\end{enumerate}

We show our results in Table~\ref{tab:voxcelebtable}. In all modes, our model is less than 11\% accurate at recognizing celebrities. Curiously, the accuracy \emph{drops} given both the image and the voice, suggesting that the way we fused a celebrity's image and voice together might be outside the model's training distribution. These results are significantly lower than CLIP's 86\% accuracy at classifying a person from their image. 

\begin{figure}[t]
\centering\includegraphics[height=0.64\paperheight]{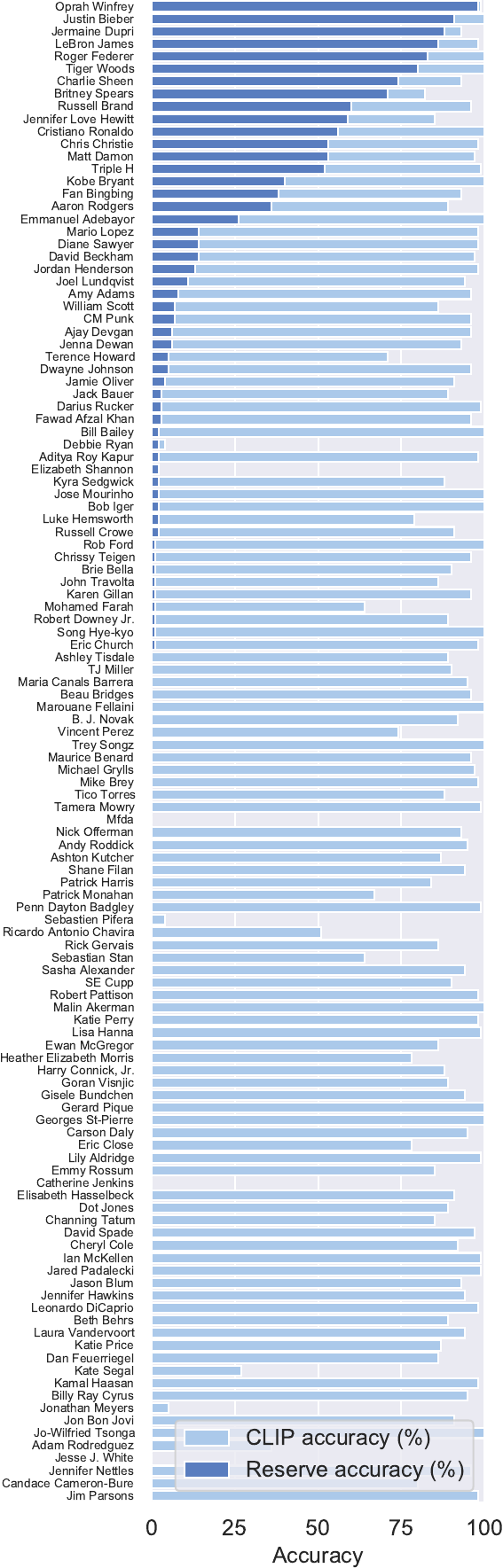}\vspace*{-2mm}
\caption{VoxCeleb2 results per-celebrity, comparing \largemodelhalfwidth~versus CLIP ViT-B/32 in the same `image-text' setting. Our model reliably recognizes A-list celebrities like Oprah Winfrey, very famous musicians (Justin Bieber) and sports players (LeBron James). However, it struggles on every other celebrity, particularly compared with CLIP. This suggests that our model primarily learns \textbf{semantic} as opposed to \textbf{recognition-level} encyclopedic knowledge.}
\label{fig:voxcelebfig}
\end{figure}

In Figure~\ref{fig:voxcelebfig}, we investigate more into which celebrities our model is best at recognizing. Only a few celebrities are reliably classified; these tend to be very famous celebrities like Oprah Winfrey and Justin Bieber. Several sports players are recognized well (including Lebron James and Roger Federer), which could imply that our model learned their identities from watching sports replays or commentary. Most other celebrities are hardly recognized, whereas CLIP does well across the board. 

\textbf{Results summary.} Together, these results show that while models like CLIP focus on encyclopedic knowledge that results in strong zero-shot person recongition accuracy, \modelname is not as effective as other models in memorizing particular celebrities-- and, thus, perhaps not as effective as memorizing particular \emph{non-celebrities}. These results suggest that \modelname's objectives and data might make it \emph{less of a concern} to release privacy-wise, versus models trained on web images with captions.

As the rest of the paper emphasizes however, \modelname performs well on tasks with temporal understanding and commonsense reasoning as the primary goal. On a broader level, these results suggest that it is possible to learn strong models about temporal reasoning \emph{without} person-level memorization, though more work is needed.

\subsection{Biases in (pre)training data.}
The `knowledge' that our model learns should be viewed as situated within YouTube \cite{keyes2021truth}, which has numerous biases (that we will discuss next). Past work has made similar observations for language model pretraining on the open web\cite{bender2021dangers}. One of the root causes of such bias is learning objectives that encourage memorization of surface level cooccurences, rather than truly causal factors \cite{haraway1988situated, bender-koller-2020-climbing,  waseem2021disembodied}. Though it is possible that in the very long term, a paradigm of \emph{grounded learning} might resolve some of these issues, the objectives in this work still likely reify biases that exist in the YouTube data.

\textbf{Platform biases.} Unlike many other pretraining efforts, that scrape data from the open internet (e.g. \cite{raffel2020t5, brown2020language, radford2021learning}) which directly leads to toxic biases (e.g. \cite{gehman2020realtoxicityprompts, dodge2021c4, birhane2021multimodal}); we trained our model on YouTube, which is a moderated platform \cite{srnicek2017platform}. Though the content moderation might perhaps reduce overtly `toxic' content, social media platforms like YouTube still contain harmful microagressions \cite{breitfeller2019finding}, and alt-lite to alt-right content \cite{ribeiro2020auditing}. Additionally, it should be mentioned that the content moderation on YouTube disproportionately filters out minoritized voices \cite{gillespie2020content}. Thus, despite us not using any word-based `blocklist,' our model's pretraining data is still biased \cite{dodge2021c4}. Even without videos being explicitly removed, the `YouTube algorithm' incentivizes the production of certain types of content over others \cite{bishop2018anxiety, strangelove2020watching}; e.g. people's roles in YouTube videos tend to be highly gendered \cite{molyneaux2008exploring}, which might bias situation understanding \cite{zhao2017men}.

\textbf{Bias amplification.}
In this work, we pretrained a model primarily on ASR text, which is itself produced by another model. The automatic captions in YouTube are known to suffer from gender bias \cite{tatman2017gender}, which our model (like neural models generally) might in turn \emph{amplify} \cite{zhao2017men}. The transcriptions on YouTube are also likely poor at handling important identity markers, like pronouns. Already, text-only models like BERT struggle with pronouns like they/them and zi/zir; our reliance on ASR text makes our corpus likely worse in this regard \cite{dev2021harms}. While past work, namely MERLOT \cite{zellers2021merlot}, `cleaned' this ASR text -- through another large language model -- we opted not to do so for this work due to computational expense. Though in that work, the ASR-denoisification was found to boost performance in VCR, it seems unlikely that it would solve this core issue of model bias.

\subsection{Dual use.} Learning connections between video, audio, and text -- though an important area of study as we have argued -- can be used for undesirable applications, beyond what we have outlined under `biases.' We outline and discuss a few below.

\textbf{Generating fake content.} A concern for pretrained models is that they can generate fake content, that could be used by `bad' actors for their ends \cite{zellers2019grover}. It should be noted that our model cannot explicitly `generate' text, audio, or vision in a direct sense. Nonetheless, however, it is possible that a finetuned or expanded version of this model could be used for that purpose -- and that our model would be \emph{more helpful} to such an actor versus them training their own (perhaps larger) model from scratch.

\textbf{Surveillance.} Our model might contain representations that enable it to be used in surveillance applications. As we note in Appendix~\ref{supp_sssec:celeb}, our model's low performance on person recognition suggests that it might perform poorly recognition-focused applications. Still, one possibility is that a \emph{neural script knowledge} could `summarize' surveillance videos in some form (like identifying an activity of interest), without identifying the person(s).

We suspect (but cannot definitively prove) that the \emph{reporting bias} of the YouTube data that it was trained on might make it poor for such a surveillance-focused task \cite{gordon2013reporting}. Namely, most surveillance videos are sparse in nature -- finding an activity of interest is like finding a `needle in a haystack' \cite{pritch2009clustered}. Though, some surveillance videos are inevitably posted on YouTube and then captioned, these disproportionately contain \emph{interesting events} (like somebody's car crashing into a house). It is not clear whether our system could be easily adapted to such a sparse problem; the amount of work required suggests that it might be out-of-scope at least for low-skill actors. On the other hand, this broad research agenda, and perhaps all of computer vision for that matter, might enable large actors to do just that \cite{zuboff2015big}; which might not be addressable through purely technical solutions \cite{green2019good}.

\textbf{Harmful outcomes if deployed.} Beyond the \emph{biases} that our system possesses, some applications of our system -- if deployed in production -- could cause harm, particularly to groups already harmed by AI systems. Of note, linking someone's voice with their appearance is not always a good thing \cite{rajunov2019nonbinary}. Likely some of the key features that our model learns -- though we did not teach it this explicitly -- involve recognizing gender, and this is harmful especially to transgender individuals \cite{hamidi2018gender}. 

\subsection{Energy consumption.}

Our model cost a lot amount of energy to pretrain \cite{strubell2019energy}; roughly 3 weeks of time on a TPU v3-512. The total carbon footprint of our work was a net 8.23 tons of CO$_2$ equivalent, which is roughly 4.5\% of the emissions of a jet plane flying round-trip from San Francisco to New York.\footnote{\textbf{CO$_2$ Calculation.} It is also important to consider the \emph{location} where these TPUs are located, as the renewables portion at each datacenter is not equal \cite{patterson2021carbon}. Our TPUs were in the `europe-west4' region, which uses on average 60\% carbon-free energy, and a Grid Carbon intensity of 0.410 kgCO$_2$eq / kWh. A single TPU v3 processor (with 8 cores over 2 chips) has a power average of 283 W, so after performing the math from \cite{patterson2021carbon}, our training cost 20,000 kWh. This gives us a net 8.23 tons of CO$_2$ equivalent. It should be mentioned that this figure only covers the \emph{electricity usage} given the chips (and the datacenter), not the raw materials involved in making these chips (which is significant \cite{villard2015drawing}).}

At the same time, it is possible that our model could save energy overall, when shared with researchers who build off of our system. Indeed, \basemodel uses less energy than MERLOT \cite{zellers2021merlot} (due to a smaller vision backbone, and smaller image sizes), MERLOT in turn is more efficient than past work which used expensive detector-based backbones (e.g. \cite{tan2019lxmert, chen2019uniter, zhang2021vinvl}), that are made more expensive because some of their computational primitives (like non-maximum suppression) are difficult to make efficient on-device.

\subsection{Synthesis.}
With these risks in mind, we release our video IDs, as well as \modelname's checkpoints, exclusively for research use. We believe that at this point in time, we as a field lack full knowledge of the privacy, bias, and dual-use risks of video-based models -- though, we hope that our analysis in this section provides a good starting point. 
For instance, while the objectives that we have studied were designed to promote learning general neural script knowledge above \emph{encyclopedic memorization}, they have not yet been tested in all possible cases.
By opening our models to the research community, we hope to promote fundamental work in uncovering both promising aspects of these systems, alongside examining their risks. We hope to contribute to these lines of research as well.

\section{Model implementation details}
\label{supp:modelimpldetails}
In this section, we discuss at a more in-depth, technical level, how we implement certain aspects of \modelname, and other details (like its runtime in FLOPs). We discuss our use of rotary position encodings (\ref{supp_ssec:rotary}), how we set the sequence lengths for the model (\ref{supp_ssec:seqlength}), measure the model's computational footprint (\ref{supp_ssec:efficiency}), list hyperparameters (\ref{supp_ssec:hyperparameters}
), and discuss several training strategies (\ref{supp_ssec:speedimprovements}.

\subsection{Rotary position encoding}
\label{supp_ssec:rotary}

We use a rotary position encoding to model the relative location of input sequences \cite{su2021roformer,rope-eleutherai}. We chose this primarily because we did not want to use absolute (additive) position embeddings, which would have to be added to the inputs of each encoder, and possibly at multiple levels in the hierarchy (e.g. for the joint encoder, the video segment index $t$ would be needed as well).

The rotary encoding uses no parameters, and instead uses a kernel trick to allow the model to recover relative distances between key and query elements in a Transformer's attention head. This can be seen as `rotating' pairs of elements; we apply the rotation to only the first \emph{half} of each 64-dimensional head, and the second half is kept as is.

\textbf{Multidimensional coordinates.} We treat each token as having a 4-dimensional position of $(h,w,\ell,t)$, corresponding to the $h,w$ coordinates in the image, the position $\ell$ in the text-sequence, and the segment index $t$. If a dimension is irrelevant to a modality (like $h,w$ for text), we set it to 0. Thus, for our various encoders, we use the following coordinate schemes:
\begin{enumerate}[rowan]
\item Video Frame Encoder (ViT): just the $h,w$ coordinates of the image; so $(h, w, 0, 0)$.
\item Audio Encoder: Only the 1-D position $\ell$ of the patch in the spectrogram: $(0, 0, \ell, 0)$.
\item Text Span Encoder: Only the 1-D position $\ell$ of the token in the input: $(0, 0, \ell, 0)$.
\item Joint encoder: Here, we use all coordinates. Inputs from the video frame encoder have coordinates $(h, w, 0, t)$, where $t$ is their segment index. The text and (pooled) audio inputs are merged, and they each have coordinates $(0, 0, \ell, t)$, where $\ell$ here is the absolute position in the entire sequence (across segments).
\end{enumerate}
As part of our implementation, we normalize the rotary coordinates. $h,w$ are scaled to be in the range $[-1/2, 1/2]$, such that text is implicitly `in the center' of the image. Likewise, $\ell$ and $t$ are scaled to be in the range of $[0,1]$. The positions are used to compute relative distances, by using a kernel trick to rotate coordinates in the keys and values of each $d_h$-sized Transformer attention head.

\subsection{Sequence lengths}
\label{supp_ssec:seqlength}
We briefly remark on the sequence lengths used by parts of the model.

\begin{enumerate}[rowan]
\item Video Frame Encoder (ViT): Most YouTube videos are widescreen (16x9). We thus used a widescreen resolution for our video frame encoder. It takes in patches of size 16x16, and we used a layout of 12 patches (in height) by 20 patches (in width). This corresponds to \textbf{192x320}. Among other factors that are important are ensuring that TPUs do not execessively pad the sequence length \cite{zhai2021scaling}. The sequence length is 241 in this case, as there is a \clstoken~token, and it gets padded to 256.

\emph{Attention pooling.} As we note in the main text, afterwards we apply attention pooling in a 2x2 grid (ignoring the \clstoken~token here). Similar to Transformer-style query,key,value attention \cite{vaswani2017attention}, the query is the average of the vectors in the 2x2 grid; the keys and values are learned projections of the vectors.
This gives us a $H/32$ by $W/32$ grid for the joint encoder (6 x 10).

\item Audio Encoder. Our model independently encodes each 1.6 second of audio (a segment has three such `subsegments'). We do this through spectrograms. Each window involves 1536 samples at a sample rate of 22500 Hz, and there are 588 samples `hops' between windows. We chose these hyperparameters largely around efficiency. We found that the Discrete Fourier Transform is fastest if the window size is close to a multiple of 2. We used a small number of mel spectrogram bins (64) because we found that at that threshold, we could reconstruct the original sequence at an acceptable level using the Griffin-Lim algorithm, \cite{griffin1984signal} which itself might be a \emph{lower bound} on quality as neural methods trained for this purpose have been shown to do better \cite{wang2017tacotron}.

In our implementation, we compute the spectrogram for an entire video segment (5 seconds) at once; this is of size $64$ mel bins by 192 windows. During pretraining, we perform what is effectively a `random crop' over the spectrogram: we extract three sequential $64 x 60$ sub-spectrograms, for each audio subsegment. We constrain them to not overlap, which means that 12 (random) windows are held out.

We note that our Audio Encoder AST is quite different from the one proposed by \cite{gong2021ast}. Though it operates over spectrograms, we opted for a linear `1-dimensional' layout rather than a two-dimensional (image-like) one. We also did not pretrain our audio encoder on any supervised data (they used ImageNet and found, perhaps surprisingly, that it helped initialize the model). We used a patch size of $64$ mel bins by $2$ windows; the resulting (1D) sequence is of size 30. After adding a \clstoken~token, the result is a sequence of length \textbf{31.}

As we note in the main text, we apply attention pooling afterwards (for all elements except the \clstoken~token), pooling by a factor of five to resize the length-30 sequence to a length of 6 `audio tokens.'

\item Text Span Encoder: We operate on spans that are at most of length 15, with an additional \clstoken~token. Its length is thus \textbf{16}.
\item Joint encoder. Let $L$ be the number of text or pooled audio tokens given to the model per segment, on average; we set $L{=}20$. Let $T$ be the number of video segments. Then, the joint model's sequence length is $T\times(L + W/32{\times}H/32)$. We set $T{=}8$ (8 video segments given to the model at a time) and used a $H{=}192$ by $W{=}320$ resolution. Our total sequence length was thus \textbf{640}.
\end{enumerate}

To better adapt our model to downstream tasks -- particularly single-image tasks like VCR \cite{zellers2019recognition}, where past work tends to use a resolution much higher than 192x320, after pretraining, we performed FixRes pretraining (for one epoch on \basemodel, and one half epoch on \largemodel \cite{touvron2019fixing}.\footnote{We had intended to do a full epoch for \largemodel, but our job got preempted, and the loss seemed to have already converged.} Here, we trained the model on larger images -- simultaneously on 288x512 widescreen images (18 patches by 32 patches), and on 384x384 square images (24 patches on each side). The joint encoder, correspondingly, uses a sequence length of 1312.

During 10 epochs of pretraining, we used a cosine decay of the learning rate down to 0.02 its maximum. During FixRes pretraining afterwards, we warmed up the learning rate to 0.02x its peak, over the first 1/5th of an epoch, and afterwards used a cosine schedule to anneal it towards 0.
\subsection{Efficiency metrics of our model}
\label{supp_ssec:efficiency}

\begin{table}[t!]
\centering
\begin{tabular}{@{} p{3cm} @{} r  @{\hspace{0.04cm}} r @{\hspace{0.2cm}}|@{\hspace{0.2cm}} r @{\hspace{0.6cm}}  p{1cm}   @{} }
& \multicolumn{3}{c}{GFlops, from} & VCR \\ \cmidrule{2-4}
Model & \parbox{1cm}{\smaller Image \newline Encoder} & \parbox{1cm}{\smaller Joint\newline Encoder} & Total & \parbox{1cm}{Q$\rightarrow$AR \newline Acc(\%) }\\[0.1cm]\toprule
UNITER-Base\cite{chen2019uniter} & 1766 & 28 & 1794 & 58.2\\
UNITER-Large\cite{chen2019uniter} & 1767 & 99 & 1867 & 62.8 \\\midrule
MERLOT \cite{zellers2021merlot}   &   236 & 67 & 303 & 65.1 \\ \midrule
\basemodelhalfwidth    &   99      &    46  &146 & 62.6 \\
\largemodelhalfwidth    &   176      &    165  &341 & \textbf{71.5} \\ 
\bottomrule
\end{tabular}
\caption{Efficiency metrics of our model versus others, measured in terms of (giga) floating point operations required to process a single image, question, and answer candidate on VCR. We compare with the overall VCR performance on the combined Q$\rightarrow$AR metric. Our \modelnamehalfwidth~family of models are significantly more efficient than prior work, with \largemodelhalfwidth~being roughly on par with MERLOT \cite{zellers2021merlot} in terms of FLOPs, yet improving accuracy by over 6\%.}
\label{tab:efficiency}
\end{table}

In Table~\ref{tab:efficiency}, we report efficiency metrics of \modelname, versus others. We calculate these metrics in the context of scoring a single VCR question and answer candidate. This requires encoding one image, and using 128 tokens for each question and answer combined (for all models). We compare against a UNITER \cite{chen2019uniter}, which is a representative VisualBERT style model, along with MERLOT \cite{zellers2021merlot}. Our models are far more efficient in terms of FLOPs, with \largemodel being roughly on par with MERLOT, yet outperforming it by 6\% in terms of VCR accuracy. We discuss key differences below:

\begin{enumerate}[rowan]
\item \textbf{UNITER}. We note that UNITER, like other VisualBERT models, uses a supervised object detection backbone \cite{Anderson2017updown}. This processes images using a ResNet 101 model \cite{he2016deep}, at a resolution of 600x800; the final ResNet `C4' block is applied densely over the entire image to obtain object-detection potentials everywhere in the image. Both factors greatly increase the FLOPs count. 

When computing UNITER's FLOPs count, we exclude operations like non-max suppression, which is an operation that is difficult to implement (and thus whose FLOP count might vary significantly depending on implementation). Our FLOPs count is thus a lower-bound. 36 detection regions are extracted, which is why the `joint encoder' for UNITER is smaller than the equivalents for MERLOT and \modelname.
\item MERLOT. This model has two key differences versus our \modelname. First, it uses a larger image resolution for VCR: 384x704, versus our 288x512. Second, it uses a hybrid ViT-ResNet50 backbone for encoding images. The backbone here is lighter weight than the object detection backbone of UNITER (in particular, the final `C4' block is removed), and thus, as shown in Table~\ref{tab:efficiency}, though it uses more FLOPs than does our \largemodel, it uses far fewer FLOPs than UNITER.
\end{enumerate}

We choose flops as our primary comparison metric as past work shows that it is one of the key factors in model scaling \cite{kaplan2020scaling, dosovitskiy2020image}. Parameters are arguably more fungible. For instance, in text-only representation learning, ALBERT \cite{lan2019albert}~demonstrates that it is possible to \emph{tie parameters} together at all layers of a BERT-like transformer, reducing parameters by an order of magnitude (while not modifying compute), with a minimal performance drop. We did not do this for this work, as we wanted to use a more `vanilla' Transformer architecture; however, it suggests that representation learning models with hundreds of millions of parameters might be \emph{FLOPs bound} as opposed to \emph{parameter-bound}.

Nonetheless, UNITER-Base has 154 million parameters, though some are frozen (86 million from their Transformer, 23 million from the word embedding layer, and then 44 million from their object detector \cite{Anderson2017updown}). UNITER-Large has 378 million parameters (303 from their Transformer, 31 million from word embeddings, and 44 million from the same object detector. Meanwhile, MERLOT has 223M parameters. Versus our \basemodel, 14 million extra parameters are due to a larger vocabulary, and 10 million parameters are due to a ResNet50 encoder -- but these parameters have a disproportionate impact in FLOPs count.

\subsection{Full model hyperparameters}
\label{supp_ssec:hyperparameters}
\newcolumntype{P}[1]{>{\centering\arraybackslash}p{#1}}
\newcommand\bothmodels[1]{\multicolumn{2}{P{4cm}}{\cellcolor{olive!5} #1}}

\begin{table}[t!]
\centering
\vspace{-2mm}
\begin{tabular}{@{} p{0.3cm} @{\hspace{0.05cm}} p{3.5cm} @{\hspace{0.1cm}} P{2cm} @{\hspace{0.1cm}} P{2cm} @{}}
\toprule
& & Base & Large \\ 
\midrule
\multirow{5}{*}{\rotatebox[origin=c]{90}{{\footnotesize Audio size}}} & Sample rate & \bothmodels{22050 Hz} \\
                                                                      & FFT hop length & \bothmodels{588 samples} \\
                                                                      & FFT window size & \bothmodels{1536} \\
                                                                      & Mel bins & \bothmodels{64} \\
                                                                      & Subsegment length & \bothmodels{60 hops, ($\approx$1.6 sec)} \\ 
                                                                      & Patch size & \bothmodels{64 mels $\times$ 2hops}\\ 
                                                                      & Pooling ratio & \bothmodels{5} \\  \midrule
                                                                      & Final size & \bothmodels{6 tokens} \\  \midrule
\multirow{4}{*}{\rotatebox[origin=c]{90}{{\footnotesize Image}}}      & ViT patch size & \bothmodels{16} \\                                                              
                                                                      & Pretraining size & \bothmodels{192 $\times$ 320} \\ 
                                                                      & Res-adaptation size & \bothmodels{288$\times$512 and 384$\times$384} \\ 
                                                                      & Pooling window & \bothmodels{2 $\times$ 2} \\  \midrule
\multirow{2}{*}{\rotatebox[origin=c]{90}{{\footnotesize Text}}}        & Max. span length & \bothmodels{15} \\                                                                
                                                                       & Mean span length & \bothmodels{5.5}  \\ \midrule
\multirow{6}{*}{\rotatebox[origin=c]{90}{{\footnotesize Joint sizes}}} & $N$ video segments & \bothmodels{16} \\
                                                                       & video segment groups & \bothmodels{2 (each with 8 segments)} \\
                                                                       & Pretraining seq. length & \bothmodels{640 (160 text\&pooled audio; 480 pooled vision)} \\
                                                                       & Res-adapted seq. length & \bothmodels{1312 (160 text\&pooled audio; 1152 pooled vision)} \\ \midrule
\multirow{5}{*}{\rotatebox[origin=c]{90}{{\footnotesize Batch sizes}}} & Videos & \bothmodels{1024} \\  
                                                                       & \# Frames (for matching) & \bothmodels{16384} \\
                                                                       & Masking rate & \bothmodels{25\% (of subsegments)} \\   
                                                                       & Text spans & \bothmodels{49152} \\      
                                                                       & Audio spans & \bothmodels{49152} \\\midrule
\multirow{7}{*}{\rotatebox[origin=c]{90}{{\footnotesize architecture}}} & Hidden size   & 768 & 1024 \\
                                                                        & Num attention heads & 12 & 16 \\
                                                                        & Size per head & \bothmodels{64}\\
                                                                        & Rotary size (per head) & \bothmodels{32}\\
                                                                        & Vision num layers & 12 & 24 \\
                                                                        & Audio num layers & \bothmodels{12}\\
                                                                        & Text-span num layers & \bothmodels{4}\\
                                                                        & Joint num layers  & 12 & 24 \\
                                                                        \midrule
\multirow{7}{*}{\rotatebox[origin=c]{90}{{\footnotesize optimizer}}} & Peak learning rate   & 4e-4 & 3e-4\\
                                                                     & Weight decay  & \bothmodels{0.1}\\
                                                                     & AdamW $\beta_2$  & \bothmodels{0.98}\\
                                                                     & AdamW $\epsilon$  & \bothmodels{1e-6}\\
                                                                     & Warmup steps & \bothmodels{3750}\\
                                                                     & Training steps & \bothmodels{750k (+ 75k for res. adaptation)}\\
                                                                     & Training epochs & \bothmodels{10 (+ 1 for res. adaptation)}\\
                                                                     \midrule
                                                                     & $\sigma$ Maximum scale & \bothmodels{100.0}\\ \midrule
& Pretraining compute & TPU v3-512 for 5 days & TPU v3-512 for 16 days \\
\end{tabular}\vspace{-2mm}
\caption{Architecture details, and pretraining hyperparameters, for both model sizes. }
\label{tab:archdetails}
\end{table}



\begin{table}[t!]
\centering
\vspace{-2mm}
\begin{tabular}{@{} p{0.3cm} @{\hspace{0.05cm}} p{3.5cm} @{\hspace{0.1cm}} P{2cm} @{\hspace{0.1cm}} P{2cm} @{}}
\toprule
& & Base & Large \\ 
\midrule
\multirow{5}{*}{\rotatebox[origin=c]{90}{{\footnotesize VCR}}} & Batch Size & \bothmodels{32} \\
                                                              & Training Epochs & \bothmodels{5} \\
                                                              & Image Size & \bothmodels{288$\times$512} \\ 
                                                              & Learning Rates Tried & {\footnotesize 1e-5, \textbf{2e-5}, 3e-5 } & {\footnotesize \textbf{8e-6}, 1e-5, 1.2e-5 } \\
                                                              & Learning Rate & 2e-5 & 8e-6 \\ \midrule
\multirow{5}{*}{\rotatebox[origin=c]{90}{{\footnotesize TVQA}}} & Batch Size & \bothmodels{32} \\
                                                              & Training Epochs & \bothmodels{3} \\
                                                              & Image Size & \bothmodels{288$\times$512} \\ 
                                                              & Learning Rates Tried & {\footnotesize \textbf{5e-6}, 1e-5} & {\footnotesize \textbf{5e-6}, 1e-5} \\
                                                              & Learning Rate & \bothmodels{5e-6} \\ \midrule
\multirow{5}{*}{\rotatebox[origin=c]{90}{{\footnotesize Kinetics-600}}} & Batch Size & \bothmodels{64} \\
                                                              & Training Epochs & \bothmodels{15} \\
                                                              & Image Size & \bothmodels{288$\times$512} \\ 
                                                              & Learning Rate & 1e-5 & 5e-6 \\
                                                              & Data Augmentation & \bothmodels{From \cite{akbari2021vatt}} \\
\end{tabular}\vspace{-2mm}

\caption{Hyperparameters for finetuning on downstream tasks. Note that for Kinetics-600, we tried to mimic VATT's setup \cite{akbari2021vatt}, including adopting their training-epoch regime and their data augmentation strategies. Our data augmentation strategies were much simpler for VCR and TVQA (random cropping, and for VCR sometimes horizontally flipping the image); we suspect that our VCR/TVQA results could be made higher if data augmentation was further explored.}
\label{tab:downstreamhp}
\end{table}

In Table~\ref{tab:archdetails}, we present full hyperparameters for our model. Among other details, we used AdamW as our optimizer, with $\beta_2=0.98$ and $\epsilon=1e-6$. We increased the learning rate linearly to its peak value (4e-4 for \basemodel, 3e-4 for \largemodel) over 3750 steps ($\frac{1}{20}$th of an epoch). Our number of warmup steps is lower than many other pretraining work; we note that all of our contrastive objectives involve learning a $\sigma$ parameter, which functions as a secondary `warmup.'

We did not use gradient clipping. We trained and evaluated in 16-bit bfloat16 precision wherever we could -- casting all gradients to that precision as well, and saving the AdamW running mean and variance to be 16-bit as well. A few times during pretraining \largemodel, we found that some values in gradients would be NaN. We addressed this by always setting NaN values to be 0. This seemed to address the symptoms of training instability -- though sometimes the training loss would spike to roughly around the same loss as random initialization, it always converged back to \emph{slightly better} than it was before the spike. We are not currently sure why this happens.

\subsection{Speed improvements during pretraining}
\label{supp_ssec:speedimprovements}

We made several high-level algorithmic and engineering implementations to our implementation, which made pretraining run faster, and that we discuss here.

\textbf{Duplicated video copies}. As mentioned in the main text, we create two copies per each video -- allowing us to learn separately \emph{how to handle audio as an input} as well as \emph{how to learn from audio}. We chose this in part because copying a video \emph{does not} increase the total compute requried by a factor of two. Instead:
\begin{enumerate}
    \item We use the image and audio encoders, to encode the underlying video frames and audio clips only once (for the two video copies), and then duplicate the encodings; this is far more efficient than encoding them both separately from scratch. 
    \item For the two video copies, we sampled two disjoint sets of masks (for which audio and text subsegments are replaced with \masktoken) at a 25\% rate. This increases the pool of negative samples for contrastive learning, again increasing training efficiency.
\end{enumerate}

\textbf{Reducing memory usage.} The memory usage of our Transformer implementation scales quadratically with sequence length, which could pose a problem since we operate on sequences of videos. We split the video into two groups of 8 segments, and encode each group separately by the joint encoder.

\textbf{Vectorization.} We vectorize all joint transformer inputs together into a single call. During this vectorization, we also encode the transcript (for the transcript-frame matching objective).

We note that this vectorization is incompatible with the Mask LM variant proposed by MERLOT \cite{zellers2021merlot}. In this variant, which the authors called `attention masking,' two transformer calls must happen \emph{sequentially} -- \textbf{first}, a language only encoder must encode the inputs and mark down (what is presumably) visually-grounded tokens; \textbf{second}, these tokens are masked for the joint encoder. We found that such an objective was unnecessary when pretraining under our contrastive span approach, which in turn enabled more efficient pretraining.

We discuss the exact pretraining data formatting technique that we used in the next section.

\section{Pretraining Data Formatting: alignment and masking}
\label{supp:pretrainingdatainfo}
In this section, we discuss how we turn a video $\mathcal{V}$ into a (masked) list of segments $\{\boldsymbol{s}_t\}$ for pretraining. 

Recall that each segment contains a video frame $\boldsymbol{v}_t$, ASR tokens $\boldsymbol{w}_t$, and audio $\boldsymbol{a}_t$. We generate the list of segments by iterating through the video with a 5-second sliding window.\footnote{Sometimes there are long `pauses' in videos where nothing gets said. When this happens -- if two segments in a row have fewer than 8 BPE tokens -- we merge them 90\% of the time, in effect `fast-forwarding' the audio and still extracting a frame from the middle. We do this at most twice, so the total length is at most 15 seconds here (in effect, a `playback' rate of 1x, 2x, or 3x). In roughly 90\% of cases, the segments are 5 seconds of length.} 

\textbf{Audio and text subsegments for masking.} We want audio to be used in part as a \emph{target} for contrastive prediction. However, during early exploration we found that 5 seconds of audio could correspond to many BPE tokens; roughly 15 on average. We use past work in language modeling as a guide \cite{joshi2020spanbert, raffel2020t5} and wanted an average span length of around 5 tokens. To get this, we split each audio segment into three equal \emph{subsegments}, each with a duration of 1.66 seconds. We can then perform masked language modeling at the \emph{aligned subsegment level}, where we mask out the text corresponding to an audio subsegment, and have the model (contrastively) predict the masked-out span of text, as well as the corresponding span of audio. We use a masking rate of 25\%, which means that a quarter of the subsegments will be corrupted and replaced by a \masktoken~token.

In theory, splitting the videos into (masked) segments ought to be straightforward. However, the key challenge that we ran into is that \textbf{the YouTube caption timing information is unreliable}. Problems might arise when we perform pretraining with both audio and text, on misaligned data. Suppose the model is given audio in segment $\boldsymbol{s}_{t-1}$ that ends with somebody saying the word `pasta.' If the alignment between audio and text is off, the model might be able to cheat the desired task by simply predicting the word `pasta' for segment $\boldsymbol{s}_t$ -- thereby turning the challenging masked-prediction task into an easier speech recognition task; we discuss this in more detail in Appendix~\ref{supp_ssec:alignmenterror}.

One way of addressing the timing issue would be to run our own ASR model over all videos, but we chose not to do this due to computational expense. Instead, we adopted two complementary strategies. First, we trained a lighweight regressor to refine the timing information (\ref{supp_ssec:aligningtime}); second, we mask audio and text conservatively, to minimize alignment errors (\ref{supp_ssec:worstcasemasking}). Finally, we discuss how we combine everything efficiently (in a vectorized way) in \ref{supp_ssec:puttingitalltogether}.

\subsection{YouTube Caption Timings}
\label{supp_ssec:alignmenterror}

YouTube provides automatically generated captions for accessibility purposes, which include timing information on each word. In the subtitle encoding that we used ({\tt vtt}), each word $w$ contains a single timestamp $t$ which corresponds to when the word should flash on-screen. The timings are \emph{mostly} accurate, but we found two key issues:

\begin{enumerate}[rowan]
\item First, they show up on average roughly 0.1 seconds before each word is spoken, which we suspect might be for usability purposes (perhaps so that while the viewer is reading the caption, they hear the word).
\item Second, with a single timestamp $t$ for each word, it is difficult to infer about pauses. For each word $w$, we can use its timestamp $t$, and the timestamps of adjacent words, to loosely infer an interval $[t_s', t_e']$ around when the word is said. However, the interval is not tight. We can only infer that the word is being actively spoken for some subinterval $[t_s, t_e]$ such that $t_s' \le t_s \le t_e \le t_e'$.\footnote{Note that this is compounded with the first problem, the ground truth interval $[t_s, t_e]$ might not be fully contained in the provided interval $[t_s', t_e']$ due to `captions being shown before audio', the error here is typically small though (0.1 seconds).}

This can lead to high absolute error (in terms of a difference between timesteps), when pauses occur. For example, suppose a speaker says a word, and then pauses. The interval given by the subtitles, $[t_s', t_e']$, might be rather large (possibly a few seconds), even though the actual word was spoken for a fraction of that time.  
\end{enumerate}

\subsection{Refining timing information}
\label{supp_ssec:aligningtime}

We trained a simple multilayer perceptron regressor to correct the timing information of YouTube transcripts. For data, we used 2000 videos with transcripts from YT-Temporal-180M, and also used Google Cloud's (highest quality, paid) ASR service to transcribe them. After aligning the words for these transcripts, this gave us tuples of the YouTube ASR word $\boldsymbol{w}$, its provided interval $[t_s', t_e']$, and the `ground truth' interval $[t_s, t_e]$.\footnote{When matching YouTube ASR to Google Cloud's ASR, we skipped words without an 'exact-match' alignment, as well as words that were over 0.25 seconds apart (i.e., where either $\delta_s > 0.25$ or $\delta_e > 0.25$} Our modeling objective was then to predict the desired offsets with respect to the provided interval: $\delta_s = t_s - t_s'$ and $\delta_e = t_e - t_e'$. We took a feature based approach. 

For each input $(w, t_s', t_e')$, we used as features:
\begin{enumerate}[rowan, label=\textbf{\roman*.}]
    \item the length of $w$ in characters,
    \item the length of $w$ in BPE tokens,
    \item whether $w$ is uppercase or not,
    \item the number of vowels in $w$,
    \item the number of punctuation characters in $w$,
    \item the value of $t_e' - t_s'$.
\end{enumerate}
We provided these features as input to the model, as well as the corresponding features for the next word, and the previous word. We z-normalized all features and used a two-layer multilayer perceptron, with a hidden size of 32 and RELU activations. We used a tanh activation at the end to bound the regression. The final predictions for $\delta_s$ (analogously for $\delta_e$) were then given by the following equation:
\begin{equation}
    \delta_s = c \tanh(\mathbf{w}\cdot \textbf{h} + b_1) + b_2 
\end{equation}
where $\textbf{h}$ is the hidden state, and with learnable parameters $c$, $\mathbf{w}$, $b_1$, and $b_2$. The learned bounds mean that, no matter what the input, the model will never predict an offset of above $c + b_2$ (of which it learned for both parameters $c \approx 0.2$ and $b_2 \approx 0.11$, so the offsets can never be above 0.3 seconds). We trained our lightweight regression model using an L$^1$ loss, and used it to correct the timing on all of the transcripts.

\subsection{Handling worst-case scenarios in masking, when alignment isn't perfect}
\label{supp_ssec:worstcasemasking}

\begin{figure*}[t!]
  \centering\small
    \includegraphics[width=\textwidth]{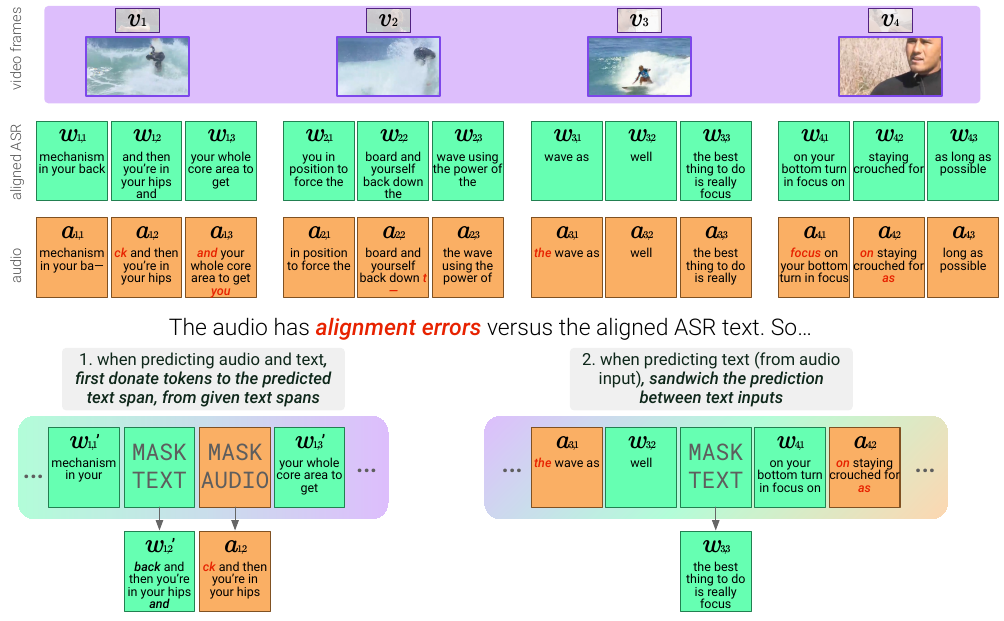}
\caption{An overview of our masking strategy for dealing with sequences of video frames, ASR, and audio. We have noisy timing information for each word, so we can align the ASR text with audio spans of 1.6 seconds each, using three sub-segments of audio and text for each video frame. However, there exist {\color{red}\textbf{alignment errors}} between the ASR and audio sub-segments -- certain words (and sub-words) have phonemes that are are in the wrong segment (like `{\color{red}\emph{\textbf{back}}}' in $\boldsymbol{w}_{1,1}$ is only partially said in the first sub-segment; the `k' sound is said in the second. When audio is only a target, we address these by `donating' tokens to predicted spans. When audio is only provided as input, we address this by sandwiching `mask' tokens between text input (so alignment does not `bleed' over).}
  \label{fig:reservemasking}
\end{figure*}
The regressor that we described reduces the average timing error of a transcript, as a preprocessing step, but it is not perfect. Thankfully, however, we find that most of the remaining alignment errors are \emph{single} words that are slightly misaligned. For instance, for three words $w_t, w_{t+1}, w_{t+2}$, the audio corresponding to the time interval around $w_t$ might contain sound from $w_{t+1}$ being spoken, but rarely $w_{t+2}$. We suspect this is primarily due to the difficulty inferring pauses: by definition, no other word can be said in a pause, so the errors are local.

We present a high level approach for masking audio and text, that in turn addresses these alignment issues (making it difficult for models to cheat). A diagram is in Figure~\ref{fig:reservemasking}.

Recall that in our framework, we only either go from `vision and text $\rightarrow$ text and audio' (\vttota), or, `vision, text, and audio $\rightarrow$ text' (\vtatot). One of the reasons we did this is to avoid allowing a model to cheat by performing speaker identification (or even `microphone identification'), which might be feasible if audio was given to the joint model as input. We can handle the two cases separately:

\begin{enumerate}[rowan]
    \item \textbf{Vision and text $\rightarrow$ text and audio} (\vttota). Here, the text as input (to the joint encoder) might overlap with the audio we are trying to predict. Our solution here is thus to \textbf{donate nearby tokens} from the predicted span, to the input. Let the span that we are trying to predict (and that we will `mask out') have a start time of $t_s$ and an ending time of $t_e$. If the final token in the previous text span, if any, has a timestamp of greater than $t_s{-}0.125$, we move it to the predicted span; likewise, if the first token in the next text span has a timestamp of less than $t_e{+}0.125$, we move it to the predicted span as well.
    \item \textbf{Vision, text, and audio $\rightarrow$ text}  (\vtatot). In this prediction task, models are given information from all modalities as input, and must predict masked-out text spans. Note that models are only given a single `speech' modality -- either text, or audio -- at each timestep. What this means is that we can carefully choose \emph{which input subsegments} to turn into `audio subsegments,' and which to turn into `text subsegments.' Our strategy is, given a masked out subsegment, to turn 80\% of adjacent subsegments into `text subsegments.'
    
    We give an illustration of this in Figure~\ref{fig:reservemasking}, part 2. Here the word `{\color{red}\emph{\textbf{focus}}}' is part of $\boldsymbol{a_{4,1}}$ but also $\boldsymbol{w_{3,3}})$. This might make $\boldsymbol{w_{3,3}})$ overly easy to predict, if we gave the model $\boldsymbol{a_{4,1}}$ as input. Our solution is thus to give the model text from $\boldsymbol{w_{3,2}})$ and from $\boldsymbol{w_{4,1}})$ as input; we are guaranteed that there is no misalignment overlap here between input and prediction spans. All of the other subsegments (not adjacent to one of the 25\% that we mask out) will be provided as audio.
\end{enumerate}

\subsection{Putting it all together, along with web text}
\label{supp_ssec:puttingitalltogether}

Finally, we discuss how we combine the various masking approaches into the prediction tasks outlined in the main text.

Each video has $N=16$ video segments, and three subsegments of audio or text spans per segment. We consider two sub-problems for this video sequence: 
\begin{enumerate}[rowanroman]
\item in \vttota, vision and text are provided as input, and the model must predict masked-out text and audio. These are done on top of separately-encoded \masktoken~tokens and \maskaudiotoken~tokens, to enable the model to learn different predictions for each modality over two separate transformer `columns.'
\item In \vtatot, vision, text and audio are provided as input, and models must predict masked-out text. Here, we use the term `predict' as a shorthand for our contrastive objective -- in which a model must match a context (a jointly-encoded \masktoken) to the exact missing span in question, where many negative contexts and spans are provided.
\end{enumerate}

We use a masking rate of 25\% for audio and text subsegments, and there are 3 subsegments per segment. This means that a single video instance gives us $48 \times 0.25{=}12$ masked-out spans of text, for each of \vttota~and \vtatot, so 24 in total (as we use disjoint masked-out subsegments). Likewise, it gives us 12 masked-out spans of audio. If we scaled these to the whole batch of 1024 videos, we would have 12k audio span options and 24k text span options. This might suffice, but scaling up the pool of candidates boosts performance in a contrastive setting, as suggested from prior work (e.g. \cite{radford2021learning}), and as our ablations (Table~\ref{tab:ablations}) support as well. Thus, we do the following:

\begin{enumerate}[rowan]
    \item \textbf{Text candidates.} We scale up the text candidates by simultaneously training the model on web text, from The Pile \cite{gao2020pile}. The joint encoder -- which can handle pooled video, pooled audio, and BPE-encoded text -- is simultaneously given a sequence of web text, for each video that we have. By performing the span-contrastive objective with this piece of web text as well, we can not only teach the model about written (as opposed to spoken) language, but we can scale up the set of text candidates as well.
    
    Let each web-text sequence be of length $L$. We first divide it into fake regions that `look like' the text subsegments in length. We do this by calculating the empirical length distribution of the text subsegments, and then using this (categorical) distribution to sample a sequence of sub-segment lengths $\ell_1, \ldots, \ell_K$.\footnote{The empirical distribution for each length, in order from a length of 1 to 15, is [0.03, 0.05, 0.08, 0.11, 0.13, 0.13, 0.12, 0.10, 0.07, 0.05, 0.03, 0.02, 0.01, 0.006, 0.003].} We clip the sampled sequence, such that $\sum_i \ell_i = L$. 
        
    Next, we mask the fake subsegments. During pretraining, we use text sequences of length $L=800$, but a model sequence length of only 640. Because we are masking spans and not individual tokens, the text sequences `shrink' when we mask them. We extract exactly 38 masked-out spans, which corresponds to around 25\% of total text.
    
    Finally, we combine the target spans that we took from the webtext sequence, with the target spans from the video. We note that sometimes -- especially in a video -- text spans might be empty. Not every 1.6 second slice of a video has someone speaking. We thus try to not use these empty spans in our contrastive objective. For each video (which is paired with text for implementation reasons) we select the `best' 48 text spans out of the (38+24) options -- penalizing empty spans, and choosing spans from videos 4x as often.
    
    These `best 48' text spans, as well as the pooled contexts that they were paired with, will be used in the contrastive objective. Aggregating over the entire batch of 1024 videos (and 1024 web text sequences), this gives us 49152 text spans as candidates, for the all-pairs symmetric softmax between text spans and contexts.
    
    \item \textbf{Audio candidates.} For each video, we note that we have exactly 12 pooled \maskaudiotoken~tokens, where the model is trying to predict the corresponding audio span. One option would be be to just use those 12 corresponding audio spans as the targets, aggregate these over the batch, and do a symmetric-cross-entropy loss.
    
    However, we can do even better \emph{for free.} Note that for the \vtatot~direction, we might have to encode many of the audio spans \emph{anyways}, using the lower level audio encoder (which simultaneously extracts a \clstoken~representation and a sequence-level pooled representation). To simplify implementation, we encode \emph{all 48 audio spans} per video. We can use these audio spans as candidates.
    
    Thus, we do the following when computing the loss over audio prediction. We aggregate all 12288 \emph{contexts} from the \maskaudiotoken~tokens in the batch, and we aggregate all 49152 candidate audio spans. We perform an all-pairs dot product between these two sets, and use it to compute a symmetric cross-entropy loss over both directions. We did not encounter any trouble using the same temperature for both directions (even though for one direction, there are 12288 options, and for the other, there are 49152). 
\end{enumerate}

The combination of these design decisions provide more `hard negatives' for the model during training. We also found that they worked well to reduce wasted computation on a TPU. For each video, the joint transformer uses one $L=640$ length sequence for transcript-frame matching, two length-$L$ sequences for the \vttota~direction (as we break it up into two groups of 8 frames each), two length $L$ sequences for the \vtatot~direction, and finally one length-$L$ sequence of text. These sequences can all be vectorized together, and the total batch size is $6\times$ the number of videos. This is helpful because using an even-numbered batch size reduces wasted computation on a TPU.

\section{Downstream Task Implementation Details}
\label{supp:downstream}
In this section, we present information for how we adapted \modelname~on downstream tasks.

\subsection{Setup for finetuned tasks}
For adapting \modelname~in a finetuned setting, we take the following approach. We use a linear warmup of the learning rate over the first half of the first epoch, with a linear decay thereafter to 0. To find the learning rate, we did a small grid search generally centered around 1e-5. Our full hyperparameters are shown in Table~\ref{tab:archdetails}.

When finetuning (and pretraining), we did not use any dropout to make implementation simpler. Instead, as a way to apply regularization, we used the same $L_2$ penalty as in pretraining (a weight decay of 0.1), but with respect to the \emph{pretrained} weights. This idea was used in \cite{wiese2017neural} among other works, and although it often tends to underperform dropout \cite{lee2019mixout}, it is simple to implement. 

\subsubsection{Visual Commonsense Reasoning}
\label{supp_ssec:vcrdetails}
As mentioned in the main text, VCR considers two subtasks: $Q{\rightarrow}A$, where models are given a question and must choose the right answer given four options; and $QA{\rightarrow}R$, where models are given a question (and the right answer) and must select the right rationale. 

In our setup for this task, we treat it as a four-way classification problem, extracting a single score from each answer or rationale candidate. An example $Q{\rightarrow}A$ is:

{\tt\tiny What is going to happen next? answer: person2 is going to say how cute person4's children are. MASK
}

An example $QA{\rightarrow}R$:

{\tt\tiny What is going to happen next? person2 is going to say how cute person4's children are. rationale: It looks like person4 is showing the photo to person2, and person2 will want to be polite. MASK
}

We extract representations from the \masktoken~position (which are of dimension $d_h$), score them with a newly-initialized $d_h \times 1$ weight matrix, and optimize scores with softmax-cross entropy.

Both VCR subtasks use only a single image. We also followed past work in `drawing on' the provided detection tags to the image \cite{zellers2021merlot}. These are unambiguous references to entities that are then referred to in the question, answer, and rationale. For example, text might reference a `{\tt person1}', which corresponds to an image region. When drawing on these detection tags, we do so in a deterministic way -- for example, `{\tt person1}' always gets the same box color. We determine the box color by hashing the object's ID (in this case, `person1') and using that to determine the hue. The model learns the connection between boxes with different hues, and the names, during finetuning.

We randomly flip images left or right, so long as there is no instance of the word `left' or `right' in the question, answer, or rationale candidates. We did no other data augmentation (other than randomly resizing images to between 100\% to 110\% of the network's size).

\subsubsection{TVQA}
\label{supp_ssec:tvqadetails}
TVQA provides models with a video, a question, and five answer candidates; we represent this as five distinct sequences for the model to score (one per candidate). The version of TVQA that we used also gives models annotations for the \emph{time region} in the video that is being referenced. It is not clear that only using this region would provide enough context to be able to understand what is going on -- enough to answer correctly. Thus, for each question, we extract 35 seconds of video around the provided time region. We then provided the model with two numbers corresponding to the time region, relative to the cropped time interval. For example, if the provided timestamp annotation is $[t_0, t_1]$, we use the following region:
\begin{align}
    t_c &= \frac{(t_0 + t_1)}{2} \\
    t_s &= t_c - 17.5 \\
    t_e &= t_c + 17.5
\end{align}
The location of $[t_0, t_1]$ in relative coordinates is then:
\begin{align}
    t_0^{r} &= \frac{t_0 - t_s}{t_e - t_s} \\
    t_1^{r} &= \frac{t_1 - t_s}{t_e - t_s}
\end{align}
We provide models with $t_0^r$ and $t_1^r$, multiplied by 100 and casted to an integer. Thus, an example TVQA instance might look like:

{\tt\small 1 to 28 What is Janice Holding on to after Chandler sends Joey to his room? Chandler's tie. \masktoken [subtitles or audio]}

This text input corresponds to the first `segment' of a video; to it we append subtitles (or audio representations) from seven segments from the provided TVQA video (with accompanying frames).

\subsubsection{Kinetics-600}
\label{supp_ssec:sdetailsvcrdetails}
We evaluate \modelname~on Activity Recognition over the Kinetics-600 dataset \cite{carreira2018short}. Here, the model has to classify a short 10-second video clip into a mutually-exclusive set of 600 categories, like `assembling bicycle' or `alligator wrestling'. 
We consider performing this task in a finetuned setting, so as to better compare to prior work. We format each example by extracting 4 video frames from the clip (sampled uniformly), and extracting 6 audio subsegments (totalling 10 seconds of audio). The model processes these inputs along with a \masktoken~token, where we extract a vector representation. We initialize the 600-way classification layer with the activations of our Text Span Encoder, over the names of the 600 categories.

We finetune the model jointly over two settings: a setting where audio is provided, and a setting where no audio is provided, to allow us to investigate both settings. We tried to closely follow VATT's finetuning approach \cite{akbari2021vatt}, including their exact data augmentation settings. We used a batch size of 64 videos (that we process simultaneously `with audio' and `without audio'). We used the same image augmentation code as VATT \cite{akbari2021vatt}, and finetuned for 15 epochs. We used a learning rate of 5e-6 for \largemodel~and 1e-5 for \basemodel.

\subsection{Setup and prompting for Zero-shot tasks}
\label{supp_ssec:zeroshotsetup}

Here, we discuss how we set up various tasks for \modelname~in a fully zero-shot setting. In addition to evaluating \modelname, we also evaluate CLIP \cite{radford2021learning} in the same zero-shot setting. CLIP is not pretrained on videos, and it cannot jointly encode text. For each task, we construct CLIP's label space by taking our prompt and substituting in each possible answer option. We average together the logits over all frames, and take a softmax, giving us a distribution over the task-specific label space.

\subsubsection{Zero-shot Action Anticipation on EPIC-Kitchens}
We study the task of action anticipation from the EPIC-Kitchens dataset \cite{Damen2021RESCALING}, a large egocentric video dataset with 700 unscripted and untrimmed videos of cooking activities. In action anticipation, a model must predict a \emph{future action} that comes $\tau_a$ seconds after a given video clip. The observed segments are of arbitrary length; we follow prior work \cite{Damen2021RESCALING} and set $\tau_a = 1$. 

The model tries to choose the correct \emph{noun} and \emph{verb} that happens next, given a list of predefined options for each. We report results on each category using the class-mean top-5 recall.

\textbf{Zero-shot inference approach.} 
We directly evaluate the pretrained \modelname~on action anticipation to verify the knowledge learned during pre-training. All prior work reported on the official leaderboard use supervision from the in-domain training set, which we do not use at all \cite{girdhar2021anticipative, furnari2020rulstm}.

For each action segment, we sample at most $N=8$ image frames and their associated audio, with fixed time interval $t=2.0$ preceding it and ending $\tau_a$ seconds before the start of the action. We append a \masktoken~token as the sole text input (at the last frame, after audio is optionally included).\footnote{We were unable to find a better text based prompt than this, as we found that they often biased the model towards linguistically relevant words; however, we suspect that such a prompt does exist.} We create short phrases out of all candidate nouns and verbs, and use that as our label space to simultaneously predict them both. We compute the score for each verb and noun independently by averaging their scores, over all labels for which they appear.

\begin{table}[t] \footnotesize
\setlength\tabcolsep{2pt}
\resizebox{\textwidth}{!}{
\begin{tabular}{clccccccccc}
& &\multicolumn{3}{c}{Overall} &\multicolumn{3}{c}{Unseen Kitchen} &\multicolumn{3}{c}{Tail Classes} \\
\cmidrule(r){3-5}
\cmidrule(r){6-8}
\cmidrule(r){9-11}
& Model & Verb & Noun & Act & Verb & Noun & Act & Verb & Noun & Act \\ 
\toprule
\multirow{12}{*}{\rotatebox[origin=c]{90}{{\footnotesize Validation }}} 
& RULSTM \cite{furnari2020rulstm}      & 27.8  & 30.8 & 14.0 & \textbf{28.8} & \textbf{27.2} & \textbf{14.2} & 19.8 & 22.0 & 11.1 \\
& AVT+ (TSN) \cite{girdhar2021anticipative} & 25.5 & 31.8 & 14.8 & 25.5 & 23.6 & 11.5 & 18.5 & 25.8 & 12.6 \\
& AVT+ \cite{girdhar2021anticipative}       & \textbf{28.2} & \textbf{32.0}  & \textbf{15.9} & 19.5 & 23.9 & 11.9 &\textbf{ 21.1} & \textbf{25.8} & \textbf{14.1} \\ 
\cmidrule(r){2-11}
& Chance & 6.4 & 2.0 & 0.2 & 14.4 & 2.9 & 0.5 & 1.6 & 0.2 & 0.1 \\
& CLIP (VIT-B/16) \cite{radford2021learning} & 13.3 & 14.5 & 2.0 & 12.3 & 8.4 & 2.1 & 14.3 & 14.3 & 1.7 \\
& CLIP (RN50x16) \cite{radford2021learning} & 16.5 & 12.8 & 2.2 & 13.4 & 7.0 & 1.2 & 17.1 & 12.6 & 2.5 \\
\cmidrule(r){2-11}
& \basemodelhalfwidth &  17.9 & 15.6 & 2.7 & 11.0 & 15.7 & 4.4 & 18.0 & 12.7 & 2.0 \\
& \largemodelhalfwidth & 15.6 & 19.3 & 4.5 & 14.1 & 18.4 & 3.4 &  14.7 & 18.5 & 4.4 \\
& \basemodelhalfwidth (+audio) & 20.9 & 17.5 & 3.7 & 15.5 & 20.1 & 4.3 & 20.7 & 14.5 & 3.2\\
& \largemodelhalfwidth (+audio) & \textbf{23.2} & \textbf{23.7} & \textbf{4.8} & \textbf{20.3} & \textbf{21.0} &\textbf{ 5.9} & \textbf{22.7} & \textbf{21.6} & \textbf{4.0}\\
\midrule
\multirow{3}{*}{\rotatebox[origin=c]{90}{{\footnotesize Test}}}
& RULSTM \cite{furnari2020rulstm}      & 25.3 & 26.7 & 11.2 & 19.4 & 26.9 & \textbf{9.7} & 17.6 & 16.0 & 7.9\\
& AVT+ \cite{girdhar2021anticipative}       & \textbf{25.6} & \textbf{28.8}  & \textbf{12.6} & 20.9 & 22.3 & 8.8 & 19.0 & 22.0 & \textbf{10.1}\\ 
\cmidrule(r){2-11}
& \largemodelhalfwidth (+audio) & 24.0 & 25.5 & 5.8 & \textbf{22.7} & \textbf{26.4} & 7.0 & \textbf{23.7} & \textbf{24.2} & 4.7 \\
\bottomrule
\end{tabular}
\caption{\modelnamehalfwidth gets competitive results on EPIC Kitchen Action Anticipation challenge with zero-shot, over methods from prior work. \label{tab:ek100table_full} }
\label{tab:ek100table}
}
\end{table}

\textbf{Results.} We show the full zero-shot action anticipation results in Table~\ref{tab:ek100table_full}. We also show our results on the test set here for our best performing model (\largemodel, with audio provided). It gets competitive results on verb and noun prediction -- with only 1.6\% and 3.3\% lower compared to the challenge winner method AVT+ \cite{girdhar2021anticipative}, which is fully supervised and use additional object-level annotations. On Unseen Kitchen and Tail Classes, our model \textbf{outperforms AVT+} on noun and verb. Overall, audio significantly improves the results -- \largemodel (+audio) outperforms \largemodel with an average 3.0\%, which suggests that it is useful for this task.

\subsubsection{Zero-shot Situated Reasoning}
Next, we evaluate on situated reasoning (STAR) \cite{wu2021star} which requires the model to capture the knowledge from surrounding situations and perform reasoning accordingly. STAR dataset includes four types of questions, including interaction, sequence, prediction, and feasibility. A model is given a video clip, a templated question, and 4 answer choices. 

\textbf{Zero-shot inference approach.} For each video clip, we sample $N=8$ image frames uniformly from the video, we also optionally include the video's sound. 

To reduce domain shift between YouTube data -- where people don't typically ask visual questions, and where ASR typically does not insert question marks -- we convert the question-answer pair into a statement. We did so using the question-answer templates provided by the author, with the answer replaced by a \masktoken. For example, \textit{``Q: What did the person do with the bottle? -- A: Put down.''} will be converted to \textit{``The person \masktoken~ the bottle.''}. 

We put the converted statement into the first frame and use the four candidate answers as a unique label space (that differs from example to example). Like with EPIC-Kitchens, we also evaluate how much audio can help by masking the audio inputs.  

\textbf{Results.} We show our zero-shot STAR results in Table~\ref{tab:zsresults} in the main text. Our base model outperforms all supervised prior work by 3.7\%. The model with audio performs better, with average 1.1\% improvement. Interestingly, \largemodel is worse than \basemodel, we suspect the reason is \largemodel is sensitive to grammar details. Given the previous example, we note that while `Put down' is a valid answer that might make sense both semantically and syntactically, a different answer `pick up' might be flagged by some English speakers as being ungrammatical: the instantiated template would then be `the person pick up the bottle.' We noticed instances of the larger model paying greater attention to these syntax-level details, even though they were not the focus of the task. It does suggest, however, that additional prompting (or label space augmentation) could resolve these issues and increase performance even further.

\subsubsection{Zero-shot LSMDC}
We evaluate our model on Movie Fill-in-the-Blank~\cite{lsmdc, maharaj2017dataset} task, which based on descriptive audio description for the visually impaired.
Given a movie clip and an aligned description with a blank in it, the task is to fill in the blank with the correct word. Following ~\cite{maharaj2017dataset}, we report prediction accuracy in test set of 30,354 examples from 10K movie clips.

\textbf{Zero-shot Inference approach.} We sample $N = 8$ video segments uniformly over the movie clip, and extract the audio and middle frame of each segment. We replace the `blank' token in each description with a \masktoken~token, and provide it (as text-based input) to the model at its final segment. For the other segments, we optionally provide the model with audio; for all segments, we provide the associated image frame. We use the vocabulary set in the LSMDC dataset as our label space (for what the `missing word' might be).

\textbf{Results.}
Our results are shown in Table~\ref{tab:zsresults} in the main text. Our model obtains 31\% when audio is included, which outperforms human text-only performance (30.2 \%)~\cite{maharaj2017dataset}, predicted by human annotators. A supervised LSTM obtains 34.4\% in this text-only setting \cite{maharaj2017dataset} which suggests that there is a certain textual bias in this task, which our model cannot learn (as it is zero-shot). This also suggests that state-of-the-art supervised models exploit patterns in this vocabulary distribution.

Without such an advantage, our model performs well, outperforming CLIP (2\%) by a large margin. This suggests that jointly reasoning over both the visual situation, and the linguistic context of the provided sentence, is helpful for zero-shot performance on LSMDC fill-in-the-blank.


\subsubsection{Zero-shot MSRVTTQA}
Finally, we evaluate our model on MSR VTT-QA, a question-answering task over videos \cite{xu2017video}. We provide a model with $N=8$ video segments sampled uniformly from the video clip, and extract an image from each one. For the first seven segments, we optionally include audio extracted from that point; at the last segment, we insert a converted version of the question, along with a \masktoken. We compare the similarity of that hidden state to the top 2000 most common answers, similar to past work \cite{zellers2021merlot}.

Similar to STAR, we convert the questions into statements to minimize drift away from the pretraining distribution. We use GPT3 prompted with several examples for this. Our exact prompt is the following:

{\tt\tiny Input: what is a car being driven through? \\
Output: a car is being driven through \_. \\
Input: who are running across screen? \\
Output: \_ are running across screen. \\
Input: when is a girl performing? \\
Output: a girl is performing at \_. \\
Input: what is a cartoon doing? \\
Output: a cartoon is \_. \\
Input: how many women talk in a bedroom? \\
Output: \_ women talk in a bedroom. \\
Input: what a man playing while dancing with others? \\
Output: a man is playing \_ while dancing with others. \\
Input: where is a flag hoisted? \\
Output: a flag is hoisted in \_. \\
Input: who talks to another man on the couch? \\
Output: \_ talks to another man on the couch. \\
Input: what does a teenage girl try to get at a public restroom? \\
Output: a teenage girl tries to get \_ at a public restroom. \\
Input: when do the models walk as the audience watches? \\
Output: the models walk as the audience watches at \_. \\
Input: what shows a person killing animals in a green forest? \\
Output: \_ shows a person killing animals in a green forest. \\
Input: who does a man ask to go on a date? \\
Output: a man asks \_ to go on a date. \\
Input: what are three people sitting on? \\
Output: three people are sitting on \_. \\
Input: \$\{question\} \\
Output:
}

Then, given a new question {\tt \$\{question\}}, GPT3 generates a converted output, wherein we can replace it's underscore with a \masktoken. GPT3 works well at this conversion, though sometimes it generates a sentence where inserting the `correct answer' feels gramatically strange. For example, the question `how many women talk in a bedroom?' suggests any integer might be a reasonable answer. On the other hand, `\_ women talk in a bedroom' implies that `one' is not a valid answer (since `women' is plural). We note that the errors caused by this conversion technique are specific to English grammar, and so if such a question-conversion approach was done in other languages, there could be more (or less) errors that directly result.

Our results are shown in Table~\ref{tab:zsresults}. Of note, our model through automatic question-conversion outperforms Just Ask \cite{yang2020just}, which performs an analogous (supervised-guided) question conversion on all its YouTube transcripts, before pretraining. Our model also outperforms CLIP, which cannot naturally handle dynamic situations.

\section{Dataset Collection}
\label{supp:datacollection}
In this section, we discussed how we curated data for \datasetname. We had several goals in mind. We wanted to use only public-facing data, which motivated our choice of YouTube as it is a public platform \emph{that users understand is public} \cite{kang2015my}. We wanted to use this platform to examine to what extent we can learn multimodal neural script knowledge from web data alone. 

Our data collection strategy in this work was informed by past work, notably MERLOT \cite{zellers2021merlot}. That paper found that increasing the diversity and scale of a video corpus both allowed for better learned representations. At the same time, the data collected by MERLOT (YT-Temporal-180M) has issues. Of note, the authors' scraping strategies -- to prioritize \emph{monetized content} -- also led to a lot of U.S. local news being in that corpus (roughly 30\% of all data). Local news might be problematic to learn from, particularly in that quantity, due to its numerous biases (e.g. racist coverage on `crime' \cite{gilliam1996crime, dixon2000overrepresentation,dixon2008crime, heider2014white}). Our goal was to expand the dataset in both diversity and size to 20 million videos, while having \emph{less local news} and without scraping private content.

\textbf{High level approach.} We adopt a similar dataset collection strategy as in MERLOT \cite{zellers2021merlot}. In the first phase, we identify a candidate set of videos ID to download. In the second phase, we open each video ID in YouTube and apply several filtering steps that go from inexpensive to expensive. The filtering steps allow us to exit early and possibly avoid downloading the video if the video seems unsuitable for our purpose from the title, description, and captions alone.

For a Datasheet \cite{gebru2018datasheets}, please see the MERLOT paper \cite{zellers2021merlot}.

\subsection{Candidate video IDs}
For MERLOT's \merlotdatasetname, the bulk of the video IDs were identified by applying breadth-first-search on YouTube channels from HowTo100M \cite{miech2019howto100m} and VLOG \cite{fouhey2018lifestyle}. Each channel often links to other channels, and given a channel it is inexpensive to obtain a list of all its videos using the youtube-dl Python package. 

In this paper, we considered numerous approaches to search for diverse, visually grounded videos. We ended up using an approach where we used YouTube's recommended videos algorithm to suggest similar videos to \merlotdatasetname. We went through all non-news and non-sports videos \merlotdatasetname, and opened each video up in YouTube. For each other video that YouTube recommended, we retrieved its channel ID -- giving us access to not just that video, but all other videos. This approach yielded 2 million channels, with 200 million videos among them.

\subsection{Filtering video IDs by channel}
Given this (large) list of channels, each with many videos, we took steps to filter it further. We used the python {\tt\small cld3} library to remove channels whose titles might not be in English. We then finetuned, and used, a language model to identify channels likely to have visually grounded videos, which we describe next. 

In more detail, we selected 2000 videos, and asked workers on Mechanical Turk to rate their level of groundedness, their genre, and whether they had explicit content or not. The questions we asked are shown in Figure~\ref{fig:suppturkinstr}. We annotated 2k videos under this schema, and trained a model to predict the annotations given video metadata. 

\begin{figure}
    {\FrameSep6pt
    \begin{framed}
    \small
    {\tt\small \$\{VIDEO\}}\\
    Q1. How would you describe the role of English speech in the video?
    \begin{enumerate}[wide, labelwidth=!,labelindent=0pt,noitemsep,topsep=2pt,label=\textbf{\alph*}.]
    \item This video doesn't have spoken English, or if it does, it's irrelevant to what's going on in the video.
    \item This video has English speech that describes, or adds onto, the visual content.
    \end{enumerate}
    Q2. Select at least one genres of the video:
    \begin{enumerate}[wide, labelwidth=!,labelindent=0pt,noitemsep,topsep=2pt,label=\textbf{\alph*}.]
    \item Gaming
    \item News
    \item How-to
    \item Chatting
    \item Sports
    \item Music
    \item Movies / Drama
    \item Documentary
    \item Miscellaneous
    \end{enumerate}
    Q3. Select if any of the following are true:
    \begin{enumerate}[wide, labelwidth=!,labelindent=0pt,noitemsep,topsep=2pt,label=\textbf{\alph*}.]
    \item A variety of objects are interacted with.
    \item A variety of actions are performed.
    \item A variety of scenes are performed.
    \item This video is a slideshow.
    \item This video contains racist or sexist content..
    \end{enumerate}
    \end{framed}}
    \vspace*{-3mm}\caption{Video annotation. We had workers on Mechanical Turk annotate 2000 videos in our dataset with this questionnaire, allowing us to then train a model to identify suitable channels for our purpose.}\vspace{-1mm}
    \label{fig:suppturkinstr}
\end{figure}

For model training, we used a slightly different setting to what we gave the crowdworkers. We trained a model to predict the labels, given a formatted list of 5 video titles from the same channel. During training, we made the weak-supervision assumption that all videos from a channel have exactly the same rating (as the video we annotated). This enabled us to collect 84k examples from our 2k annotations. The model we chose was T5-base model \cite{raffel2020t5}, which generates the labels left-to-right in text form (and which we converted automatically to a structured representation).

We then used this model to identify channels that seem especially promising. For each channel with at least 5 videos, we randomly sampled 8 sets of length-5 videos, and used the finetuned T5 model to classify them. We filtered out any channel that had at least 25\% of likely non-English or irrelevant-English videos, any channel that had at least 25\% of slideshows, and any channel that likely had racist or sexist content. 

One side benefit of this model is that it allowed us to estimate our videos' genre breakdown before downloading them. We found 1\% Gaming videos, 11\% News videos, 20\% How-To videos, 20\% `chatting' videos, 5\% sports videos, 5\% Music videos, 3\% Movies/Drama videos, 4\% Documentary videos, and 31\% Miscellaneous. The Gaming videos were then filtered out.

We used the classification model to create a budget for how many videos to download from each channel; with the aim to download more videos from likely more-grounded channels. Using the answers to Q3 (from Figure~\ref{fig:suppturkinstr}), we gave each channel 1 point for likely having `a variety of objects', 2 points for `a variety of actions', and 0.5 points for `a variety of scenes.' We subtracted 3 points if it was likely to be a slideshow. (Likely-racist or sexist channels were already filtered out.) We then z-normalized and softmaxed the channel scores, and used the result as the channel-level budgets. Any channel with an aggregate `interestingness' score of 1 standard deviation above the mean would then have a budget of 8x larger than the mean. We clipped the channel-level budgets to include at most 500 videos per channel.

This process (finally!) gave us 30 million YouTube video IDs that were likely to be high-quality. 

\subsection{Filtering videos from their metadata}
Last, we filtered and downloaded these videos using a filtering approach similar to \cite{zellers2021merlot}. We first retrieved the video metadata and used it to filter out `gaming' videos. We then retrieved the video's transcript, and filtered out any video without a `dense' span of spoken words -- defined as an interval of 30 seconds where at least 50 words are spoken. Additionally, we used the Python package {\small\tt cld3} to filter out any transcript with a probability of less than 80\% of being English. Last, we used a hidden feature in the YouTube API to download four thumbnails of the video. Using the image classification model from \cite{zellers2021merlot}, we filtered out videos whose four thumbnails had an average cosine similarity of above 85\%, or that contained fewer than 1 object from COCO.

Unlike \cite{zellers2021merlot}, we did not use a sequence-to-sequence model to `translate' spoken text to text that appears more stylistically like written English (i.e., by adding capitalization and punctuation, and removing filler words).

\section{Additional Experiments and Exploration}
\label{supp:addlexperiments}
In this section, we briefly include additional experiments, showcasing our model's performance on specific tasks that do not necessarily require multimodal script knowledge.

\subsection{Zero-shot Audio classification}

\begin{figure}[t!]
  \centering\small
    \includegraphics[width=0.75\textwidth]{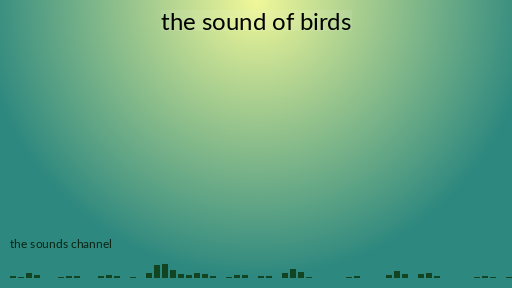}
\caption{An image prompt used in zero-shot audio classification. Here, ``the sound of'' is always inserted, and the word ``birds'' is one of the labels in ESC50 \cite{ESC50}. We consider one image prompt for each label in ESC50 (or whichever dataset we are using).}\vspace*{-2mm}
  \label{fig:image_prompt}
\end{figure}

We evaluate \modelname~on the task of zero-shot audio classification, to study to what extent its learned audio representations can directly predict text-based labels. We conduct this evaluation on environmental sounds from ESC50~\cite{ESC50}, urban sounds from US8K~\cite{US8K}, and (as part of the privacy-minded exploration in Appendix~\ref{supp:broaderimpactstatement}) celebrity voices from VoxCeleb2 \cite{nagrani2017voxceleb}.

We consider the format where we encode an audio input into a \clstoken~level representation, and retrieve the most-similar label given a set of encoded options. We encode the audio input with our encoder, which takes in as input audio clips of length at most 1.6 seconds. For shorter audio clips (like many sounds in ESC50), we repeat them in time until their length is at least 1.6 seconds. For longer audio clips, we encode multiple \clstoken~representations and then average the resulting vectors.

We consider the following ways to encode the labels:
 \begin{table}[t!]\small
	\centering
	{\setlength{\tabcolsep}{.85em}
		\makebox[\linewidth]{\resizebox{\linewidth}{!}{%
				\begin{tabular}{@{}p{2cm}p{2cm}ccc@{}}
				& & \multicolumn{3}{c}{Accuracy (\%)} \\ 
				Model & Prompting & ESC50 & US8K & VoxCeleb2 \\ \toprule
				AudioClip	&  & 68.6 & 68.8 & \\ \midrule
				\multirow{3}{*}{{\smaller \largemodelhalfwidth}}	& Text-only. & 41.6 & 60.2 & 10.8 \\
					& Image-only. & 42.8 & 54.3 & \textbf{13.3} \\
					& Image and text. & \textbf{52.2} & \textbf{62.3} & \phantom{0}9.6 \\
					\bottomrule
	\end{tabular}}}}
	\caption{\label{tab:zeroshot-audio}
		Zero-shot audio classification accuracies (\%) on ESC50~\cite{ESC50}, US8K~\cite{US8K}, and VoxCeleb2~\cite{nagrani2017voxceleb}. We compare our model with AudioClip \cite{guzhov2021audioclip}, which was pretrained on supervised data from AudioSet \cite{AudioSet}. Our \modelnamehalfwidth~performs well across the board, especially when given \emph{both the image and the text} as a prompt -- demonstrating its OCR capability.
	}
\end{table}

\begin{enumerate}[rowan]
\item \textbf{Text-only}. Inspired by the prompt `a photo of', which is used in CLIP's zero-shot image classification task \cite{radford2021learning}, we give \modelname's joint encoder a blank image, with associated tokens {\small\tt the sound of \$\{label\}}. We do this once for each label, giving us a single `target' vector for each possible label in the dataset.
\item \textbf{Image-only}. Inspired by YouTube videos of sound effects\footnote{For instance, \href{https://www.youtube.com/watch?v=VmgKryu4__k}{youtu.be/VmgKryu4\_\_k}.}, we created image-only prompts that suggest a sound (of the target class) is playing in the background. An example is shown in Figure~\ref{fig:image_prompt}. We encode each image with our joint encoder, and do this once for each label.

We note that for VoxCeleb2, we use face images of celebrities rather than this image-based prompt, due to our interest in exploring whether models can perform person-level recognition due to the privacy issue (Appendix~\ref{supp_sssec:celeb}).

\item \textbf{Image and text.} Here, we combine both of the above options: encoding one input for each label, using both the image and text prompt.
\end{enumerate}
For each prompt, we append the token `\maskaudiotoken' and extract the hidden state from there, as our final representation for that label.

\begin{figure*}
  \centering\small
    \includegraphics[width=\textwidth]{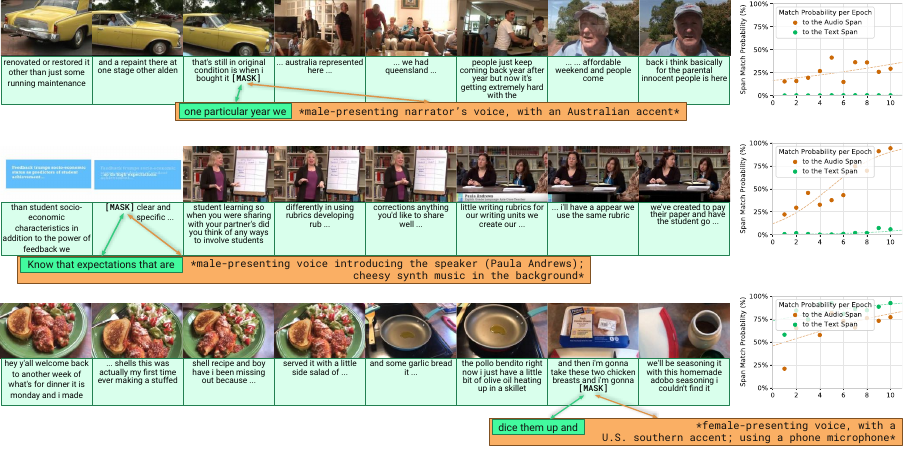}
\caption{\textbf{\masktoken ed audio self-supervision on different examples}. Similar to Figure~\ref{fig:qualfig}, we show predictions from \basemodel~over the course of pretraining. Match performance increases over time. The audio prediction in the first row is perhaps made easier by the speaker's australian accent. The audio prediction in the second row is perhaps easier due to the lecture-video setting. In the third row, both audio and text span prediction improves, with text being slightly favored in the end. This might be in part because of the truncation we do on audio (Section~\ref{supp_ssec:worstcasemasking}) -- the audio span is shorter than the text span of `dice them up and' so as to not leak information, making prediction more challenging.
}
  \label{fig:qualfigrandom}
\end{figure*}

We present our results in Table~\ref{tab:zeroshot-audio}. The results show, possibly surprisingly, that \modelname~can perform optical character recognition over image prompts like Figure~\ref{fig:image_prompt} -- given just the image, its accuracy on ESC50 is higher than given just text. Its accuracy on ESC50 and US8K improves further when given both an image and text.

These results are slightly different for VoxCeleb2, which emphasizes long-tail recognition of people -- something that might be more encyclopedic than semantic, and that we did not wish to optimize in this work. There, when given an image of a celebrity's face, it demonstrates some capacity at linking it with one of their audio clips -- a capacity that \emph{decreases} if prompted with additional text. We suspect that this is due to interpreting the given text as spoken, for example, \emph{Justin Bieber himself saying} `the sound of Justin Bieber.' On all celebrities, \modelname~struggles versus recognition-focused models like CLIP \cite{radford2021learning} (Appendix~\ref{supp_sssec:celeb}).

Overall, our model displays strong audio understanding ability. In comparison, AudioCLIP \cite{guzhov2021audioclip} (which is supervised on human-annotated labels from AudioSet \cite{AudioSet}), performs 16\% higher on ESC50, and 6.4\% higher on US8K. 

\subsection{Additional Qualitative Analysis}
In Figure~\ref{fig:qualfigrandom}, we include an additional figure of examples, of the same format as Figure~\ref{fig:qualfig}. The examples are chosen randomly -- not by how much \modelname~improved at retrieving their audio or text spans over the course of training.


\end{document}